\documentclass[lettersize,journal]{IEEEtran}
\usepackage{amsmath,amsfonts}
\usepackage{algorithmic}
\usepackage{algorithm}
\usepackage{array}
\usepackage[caption=false,font=normalsize,labelfont=sf,textfont=sf]{subfig}
\usepackage{textcomp}
\usepackage{stfloats}
\usepackage{url}
\usepackage{verbatim}
\usepackage{graphicx}
\usepackage{hyperref}
\usepackage{multirow}
\usepackage{booktabs}
\usepackage{cite}

\usepackage{amsmath}
\usepackage{amssymb}
\usepackage{mathtools}
\usepackage{amsthm}

\theoremstyle{plain}
\newtheorem{theorem}{Theorem}[section]
\newtheorem{proposition}[theorem]{Proposition}

\theoremstyle{definition}

\newtheorem{assumption}[theorem]{Assumption}
\theoremstyle{remark}

\begin{document}

\title{New Insights on Unfolding and Fine-tuning Quantum Federated Learning}

\author{Shanika Iroshi Nanayakkara, Shiva Raj Pokhrel,~\IEEEmembership{~Senior IEEE Member}
\thanks{Shanika Iroshi Nanayakkara is with School of IT,  Deakin University, VIC 3125, Burwood, Australia, (e-mail: s.nanayakkara@deakin.edu.au).
}
\thanks{Shiva Raj Pokhrel is with School of IT,  Deakin University, VIC 3125, Burwood, Australia, (e-mail: shiva.pokhrel@deakin.edu.au).
}
}

\markboth{Journal of \LaTeX\ Class Files,~Vol.~14, No.~8, June~2025}%
{Shell \MakeLowercase{\textit{et al.}}: A Sample Article Using IEEEtran.cls for IEEE Journals}


\maketitle

\begin{abstract}
Client heterogeneity poses significant challenges to the performance of Quantum Federated Learning (QFL). To overcome these limitations, we propose a new approach leveraging \textit{deep unfolding}, which enables clients to autonomously optimize hyperparameters, such as learning rates and regularization factors, based on their specific training behavior. This dynamic adaptation mitigates overfitting and ensures robust optimization in highly heterogeneous environments where standard aggregation methods often fail. Our framework achieves approximately 90\% accuracy, significantly outperforming traditional methods, which typically yield around 55\% accuracy, as demonstrated through real-time training on IBM’s quantum hardware and Qiskit Aer simulators. By developing self-adaptive \textit{fine-tuning}, the proposed method proves particularly effective in critical applications such as gene expression analysis and cancer detection, enhancing diagnostic precision and predictive modeling within quantum systems. Our results 
are attributed to convergence-aware, learnable optimization steps
intrinsic to the deep unfolded framework which maintains the generalization. Hence, this study addresses the core limitations of conventional QFL, streamlining its applicability to any complex challenges such as healthcare and genomic research.
\end{abstract}


\section{Introduction}
\begin{figure}[h]
     \centering
    \includegraphics[width=0.8\linewidth]{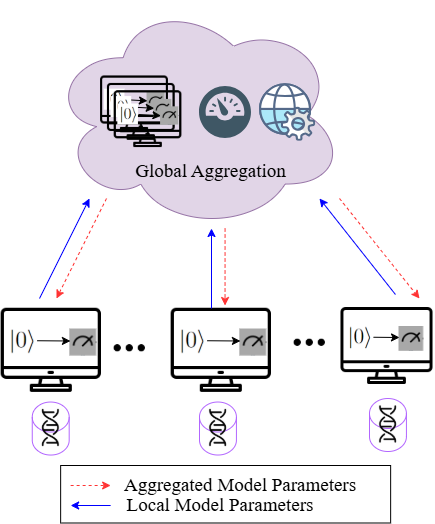}
  \caption{A High-Level View of Quantum Federated Learning Setup}
\label{fig:fl_Archi} 
\end{figure}

\IEEEPARstart{F}{ederated} learning (FL) enables decentralized model training across multiple devices while preserving data privacy by keeping local datasets intact. Quantum machine learning (QML), leveraging parameterized quantum circuits, has demonstrated strong learning performance in benchmarking tasks, validated through both numerical simulations and real quantum hardware~\cite{jerbi2024shadows}. Quantum federated learning (QFL)~\cite{gurung2024personalized}, illustrated in Figure~\ref{fig:fl_Archi}, is particularly promising for privacy-sensitive applications such as genomic data analysis, with advances in federated learning that ensure data remains local and mitigate privacy risks~\cite{chehimi2023foundations}.

\begin{table}[hbt!]
\small
\caption{Notation and Descriptions}
\label{tbl:Notation}
\vskip 0.15in
\begin{center}
\begin{small}
\begin{tabular}{@{}l|l|p{5.5cm}@{}}
\toprule
\textbf{\rotatebox{90}{Category}} & \textbf{Notation} & \textbf{Description} \\
\midrule

\multirow{7}{*}{\rotatebox{90}{Client \& Indexing}} 
& $k$ & Number of clients \\
& $i$ & Client index \\
& $j$ & Perceptron \\
& $t$ & Unfolded iteration \\
&$r$ & Federated round \\
& $l$ & Layer \\
& $S_n$ & Set of selected nodes \\
& $N_n / N_t$ & Local to total data ratio \\

\midrule
\multirow{4}{*}{\rotatebox{90}{Learning}} 
& $\eta$ & Learning rate \\
& $E$ & Tunable local epoch \\
& $\omega_i$ & Aggregation weight for client \( i \in [N] \) \\
& $\lambda.C$ & Regularization for $\sum_{i=1}^N \Theta_i \leq P_{\text{total}}$ \\

\midrule
\multirow{7}{*}{\rotatebox{90}{Gradients}} 
& $\nabla L_n$ & Gradient of the loss \\
& $\nabla_\theta F_i$ & Client \( i \)'s gradient for parameter optimization \\
& $\nabla_{U_\theta}$ & Gradient w.r.t. pole unitary $U_\theta$ \\
& $\nabla_{U_\phi}$ & Gradient w.r.t. angle unitary $U_\phi$ \\
& $\nabla_\eta \mathcal{L}_{\text{local}}, \nabla_\delta \mathcal{L}_{\text{local}}$ & Hyperparameter adaptation gradients \\
& $g_k$ & Gradient contribution \\
& $\Xi_l$ & Gradient group scaling factor \\

\midrule
\multirow{11}{*}{\rotatebox{90}{Quantum Circuit}} 
& $U$ & Unitary \\
& $U^{l,j}_{t+1}$ & Updated global unitary \\
& $U^\dagger$ & Hermitian adjoint / conjugate transpose \\
& $\theta$ & Pole parameters \\
& $\phi$ & Angle parameters \\
& $\tilde{U}_{\theta_G}$ & Global aggregated pole unitary \\
& $\tilde{U}_{\phi_G}$ & Global aggregated angle unitary \\
& $\tilde{U}_{\theta_n}, \tilde{U}_{\phi_n}$ & Local updated pole and angle unitaries \\
& $c_n^\theta, c_n^\phi$ & Participation indicators \\
& $R_z(\theta_i)$ & Rotation gate encoding \\
& $f_0$ & Frequency of measuring \\

\midrule
\multirow{2}{*}{\rotatebox{90}{Metrics}} 
& $e_m^2$ & Squared error $(y_m - f_{U_{\theta_k}}(|\psi_m\rangle))$ \\
& $M_{(best)}$ & Global mean best position \\
& $\nabla_{r}^{(t)}$ & Local attraction \\
& $L_r^{(t)}$ & Local attractor \\
& $\gamma$, $\beta$ & QPSO hyperparameters \\

\midrule
\multirow{4}{*}{\rotatebox{90}{Comm/Noise}} 
&$q_i$ & Normalized selection probability for client \\
& $q_i$ & $\frac{\tilde{q}_i^p C_i}{\sum_{j=1}^n \tilde{q}_j^p C_j}$ \\
&$C_i$ &computational capacity if i client\\
& $\tilde{q}_i$ & $D_i(|\psi\rangle\langle\psi|)$, fidelity-based quality score \\
&$D_i$ &trace-distance\\
&$|\psi\rangle$& column vector or quantum state\\

& $h_i$ & Channel gain \\
& $\sigma_2$ & Noise power \\

\bottomrule
\end{tabular}
\end{small}
\end{center}
\vskip -0.1in
\end{table}

In this paper, we propose a novel approach that integrates \textit{deep unfolding} with \textit{quantum federated learning} to enhance data modeling. Deep unfolding, which represents iterative optimization steps as layers in a neural network-like architecture, enables faster convergence, fine-tuned performance, and improved interpretability~\cite{arai2024deep, monga2021algounrolling}. By dynamically \textit{fine-tuning} parameters such as learning and perturbation rates, our method optimizes quantum models for efficient training on decentralized genomic data, pushing the boundaries of accuracy and privacy preservation. Genomic datasets are high-dimensional, noisy, and sparse. Our approach of fine-tuning and unfolding over quantum federated learning addresses key challenges in genomic data analysis, such as high computational cost, overfitting, and slow convergence, while preserving data privacy.  

\begin{table}[t]
\caption{Acronyms and descriptions.}
\label{tbl:Acronyms}
\vskip 0.15in
\begin{center}
\begin{small}
\begin{tabular}{l|p{6cm}}
\toprule
\textbf{Acronym} & \textbf{Description} \\
\midrule
QML & Quantum Machine Learning \\
QNN & Quantum Neural Networks \\
VQC & Variational Quantum Circuit \\
DQNN & Deep Unfolded Quantum Neural Network \\
SPSA & Simultaneous Perturbation Stochastic App: \\
FL & Federated Learning \\
DUN & Deep Unfolding Network \\
QRAM & Quantum Random Access Memory \\
PCA & Principal Component Analysis \\
FQNGD & Federated Quantum Natural Gradient Descent \\
IID & Independent and Identically Distributed \\
AV & Autonomous Vehicle \\
GHZ State & Greenberger-Horne-Zeilinger State \\
FT-QNN & Federated Teleportation QNN \\
QSGD & Quantum Stochastic Gradient Descent \\
OQFL & Optimized Quantum-Based Federated Learning \\
Aer\_Opt & Aer Optimization level \\
\bottomrule
\end{tabular}
\end{small}
\end{center}
\vskip -0.1in
\end{table}

Our main contributions are as follows:
\begin{itemize}
    \item We propose a novel architecture integrating deep unfolding with quantum federated learning to accelerate QFL training.
    \item We introduce autonomous fine-tuning using a `learning to learn' approach, replacing manual tuning with dynamic adaptability for better convergence and robustness.
    \item We evaluate experiments on genome sequencing and cancer datasets, showcasing high accuracy and versatility in real-world applications.
    \item We extend our simulations on both the Qiskit-aer simulator and Real IBM Quantum QPUs are demonstrating consistent high accuracy and practical applicability in real-time quantum federated learning.
\end{itemize}

\subsection{Related Work}

\begin{table*}[htb!]
\centering
\scriptsize
\caption{Summary of Quantum Federated Learning (QFL) Literature}
\begin{tabular}{|p{1cm}|p{0.5cm}|p{4.2cm}|p{4.5cm}|p{3cm}|p{0.5cm}|p{0.5cm}|p{0.5cm}|}
\hline
\textbf{\rotatebox{90}{Category}} & \textbf{Ref} & \textbf{Global Aggregation} & \textbf{Local Update} & \textbf{Parameter Tuning} & \rotatebox{90}{\textbf{L2L}} & \rotatebox{90}{\textbf{QNN+Genome}} & \rotatebox{90}{\textbf{Qiskit}} \\
\hline

\multirow{5}{*}{\rotatebox{90}{Security}} 
& \cite{li2021quantumblind} 
& $\bar{U}_{r+1} = \bar{U}_r - \eta f_{Adam}(U_r^{(k)})$
& $\nabla L(U_r^{(k)})_{clip} + \frac{2R}{\mu}N(0,1)$
& $(\eta_r)_{Adam}$ & \(\times\) & \(\times\) & \(\times\) \\

\cline{2-8}
& \cite{chu2023cryptoqfl} 
& $\sum_{i=1}^N \mathcal{U}_{\text{QOTP}} \cdot U_{\theta_i}$
& $\eta \cdot \text{De}\left(\mathcal{U}_{\text{QOTP}}^{-1} \cdot \left\{ \left|\phi_{g_i}\rangle \right| \right\}_{i=1}^N \right)$
& \text{Update}(\{x, z\}) & \(\times\) & \(\times\) & \(\checkmark\) \\

\cline{2-8}
& \cite{yun2022slimmable} 
& $U_{(\theta,\phi)} - \eta_r \nabla_{U_{(\theta,\phi)}} L(U_\phi, U_\theta)$
& $\frac{\sum_{n=1}^N c_n^{(\theta,\phi)} \tilde{U}_{(\theta_n,\phi_n)}}{\sum_{n=1}^N c_n^{(\theta,\phi)}}$
& $\frac{\eta}{1 + \alpha r}$ & \(\times\) & \(\times\) & \(\times\) \\

\cline{2-8}
& \cite{yun2022quantum} 
& $\Theta_r - \eta_r \sum_{l=1}^L \frac{1}{|X_l|} \sum_{k \in X_l} g_{k, r} \cdot \Xi_l$
& $\theta_{k}^{(r, e)} - \eta_r \nabla_{\theta_{k, r, e}} L_{k, l}^{(r, e)}$
& fixed & \(\times\) & \(\checkmark\) & \(\times\) \\

\cline{2-8}
& \cite{zhang2022federated} 
& $\sum_{i=1}^N \theta_i = \arccos(2f_0 - 1)$
& $q_i' = R_z(\theta_i) \cdot q_i$
& fixed & \(\times\) & \(\checkmark\) & \(\times\) \\

\hline

\multirow{2}{*}{\rotatebox{90}{AV}} 
& \cite{yamany2021oqfl} 
& $W^{(i)} = \sum_{r \in R} \frac{|D_r|}{|D|} W_r^{(i)}$
& $W_r - \frac{\eta}{E} \sum_{{e=1}^{E},{\theta \in \Theta}} \nabla L(\theta_r^{(r)})$
& $\gamma \cdot \Delta_r^{(r)} \cdot \ln\left(\frac{1}{\beta}\right)$ & \(\checkmark\) & \(\times\) & \(\times\) \\

\cline{2-8}
& \cite{narottama2023federated} 
& $\Theta_{\text{cloud}} \leftarrow \arg \min_{\Theta} L_{\text{global}}(\Omega)$
& $U_i^{[m]} \leftarrow \arg \min_{U_i} L_{\text{local}}^{[m]}(\Omega)$
& fixed & \(\times\) & \(\checkmark\) & \(\times\) \\

\hline

\multirow{3}{*}{\rotatebox{90}{Health}} 
& \cite{zhao2023nonQFLinf} 
& $\sum_{i=1}^n q_i\cdot M_i(|\psi\rangle\langle\psi|)$
& $e^{-i\eta \nabla L_n} U_{n,r}^{l,j}$
& $\eta_r - \alpha \frac{\partial L_{\text{global}}}{\partial \eta_r}$ & \(\times\) & \(\times\) & \(\times\) \\

\cline{2-8}
& \cite{huang2022quantum} 
& $U_{\text{shared}}^{(r)} - \eta \cdot \frac{1}{n} \sum_{i=1}^n \nabla \mathcal{L}_i^{(r)}$
& $\theta_i^{(r)} - \eta \cdot \nabla \mathcal{L}_i^{(r)}(\theta_i)$
& fixed & \(\times\) & \(\times\) & \(\checkmark\) \\

\cline{2-8}
& \cite{pokhrel2024quantum} 
& $w_g = \sum_{i=1}^m \frac{|D_i|}{|D|} w_i$
& $w_i = \text{minimize\_loss}(U_{Q_i}(X_{q_i}), y_i)$
& fixed & \(\times\) & \(\times\) & \(\checkmark\) \\

\hline

\multirow{3}{*}{\rotatebox{90}{Other}} 
& \cite{qi2023optimizingnaturalGD} 
& $\bar{U}_r - \eta \sum_{i=1}^i w_i \cdot U_r^{(i)}$
& $g_i^+(U_r^{(i)}) \frac{\partial \mathcal{L}(U_r^{(i)}; S_i)}{\partial U_r^{(i)}}$
& $\eta_r, w_i = \frac{|S_i|}{|S|}$ & \(\times\) & \(\times\) & \(\checkmark\) \\

\cline{2-8}
& \cite{chehimi2022quantum} 
& $\sum_{i \in k} \frac{M_i}{\sum_{j \in i} M_j} \cdot U_{\theta_i}^h$
& $U_{\theta_i}^h - \eta \nabla_{U_{\theta_k}} \frac{1}{2M} \sum_{m=1}^M e_m^2$
& fixed & \(\times\) & \(\times\) & \(\checkmark\) \\

\cline{2-8}
& \cite{xia2021quantumfed} 
& $\left( \prod_{k=1}^{I_l} \prod_{n \in S_n} U^{l,j}_{n,k} \right) U^{l,j}_r$
& $e^{i \epsilon \frac{N_n}{N_t} K^l_n} U^{l,j}_n$
& $\frac{\partial \mathcal{L}_n}{\partial U^{l,j}_n} \left(U^{l,j}_n\right)^\dagger$
& \(\times\) & \(\times\) & \(\times\) \\

\hline
All & \textbf{Ours}
& $(U_{r+1}^{l,j})^{Best_l} = \frac{1}{N} \sum_{i=1}^N U_{i}^{l,j}$
& $U_{i,r+1}^{l,j} = U_{i,r}^{l,j} - \eta_r \cdot \nabla U_{i,r}^{l,j}$
& $g(\eta_r, \delta_r, M_i)$ & \(\checkmark\) & \(\checkmark\) & \(\checkmark\) \\
\hline
\end{tabular}
\label{tbl:literature}
\end{table*}

Conventionally, static interval-based updates lead to stale global unitaries, 
and the lack of learnable parameters restricts adaptability \cite{xia2021quantumfed}.
Weights using density estimators $D_i$ over federated iterations 
use
global aggregation with local channels $(M_{i})$ \cite{zhao2023nonQFLinf}.
In contrast, we introduce promising adaptiveness and balanced contribution,
enabling faster convergence and better alignment with quantum dynamics.

Predefined optimization framework that utilizes fixed hyperparameters and contribution weights limits the flexibility to adapt to varying dataset distributions or circuit structures \cite{qi2023optimizingnaturalGD}.
Blind quantum computation allows a quantum client to delegate computation to a quantum server so that the computation remains hidden from the quantum server \cite{li2021quantumblind}. Herein, the local model performs the gradient computation under some cryptographic protocols (e.g. UBQC) and updates after receiving the corresponding model updates by the Adam optimizer by adding Laplace or Gaussian noise to the clipped gradients for privacy \cite{li2021quantumblind}.

SlimQFL dynamically communicates angle and pole parameters while keeping hyperparameters static and predefined \cite{yun2022slimmable}, only training on data up to October 2023. Our method gradually moves from manual-decay learning rates to a data-driven, model-adaptive learning rate for improved flexibility and efficiency.
On the other hand, formulate secure updates using quantum homomorphic encryption and send updates to a central quantum server with quantum data sets with IID and non-IID client distributions  \cite{chehimi2022quantum, chehimi2023foundations}.
Moreover, communication efficiency of QFL improved by quantizing the local gradient updates into ternary values, sending only the non-zero gradient components rather than the entire gradient range \cite{chu2023cryptoqfl}.
A VQA framework for decentralized data, ensures privacy with shared parameters aggregated via a CPU introducing QSGD to reduce gradient computation costs and combined it with Adam for adaptive learning rates, accelerating convergence \cite{huang2022quantum}.

Optimized Quantum-based Federated Learning (OQFL) integrates Quantum-Behaved Particle Swarm Optimization (QPSO) providing swarm-inspired probabilistic search dynamics for global convergence \cite{yamany2021oqfl}. In contrast, the proposed DUQFL learns the optimization process, potentially leading to task-specific, and offers a more flexible framework as hyperparameter adjustments are not pre-defined but learned iteratively, improving adaptability across heterogeneous datasets.
FT-QNN, which allows the cloud QNN to obtain the outputs of edge QNNs in NOMA-based systems using IBM Qiskit platform \cite{narottama2023federated}.

Entangled slimmable Quantum Federated Learning (eSQFL) provide entangled controlled universal (CU) gates and an in place fidelity distillation (IPFD) regularizer for high-order entanglement entropy control and inter-depth command interference of eSQNNs architecture \cite{yun2022quantum}. 
A decentralized quantum secure aggregate (QSA) protocol based on GHZ-state ensures a maximally entangled state used to correlate the qubits of all participants \cite{zhang2022federated}.

Our proposed DUQFL introduces an innovative theoretical design framework with autonomous fine-tuning through a `learning to learn' unfolding approach. This dynamic self-adaptive behaviour significantly enhances flexibility and enables superior convergence, particularly for quantum federated learning on non-IID datasets. While exiting preliminary work \cite{pokhrel2024quantum} focuses on genomic data, our approach advances the field by addressing the critical challenge of adaptive hyperparameter optimization in a theoretically robust manner.

\section{Proposed Methodology}
\label{sec-framework}

\begin{figure}[t]
     \centering
    \includegraphics[width=1.0\linewidth]{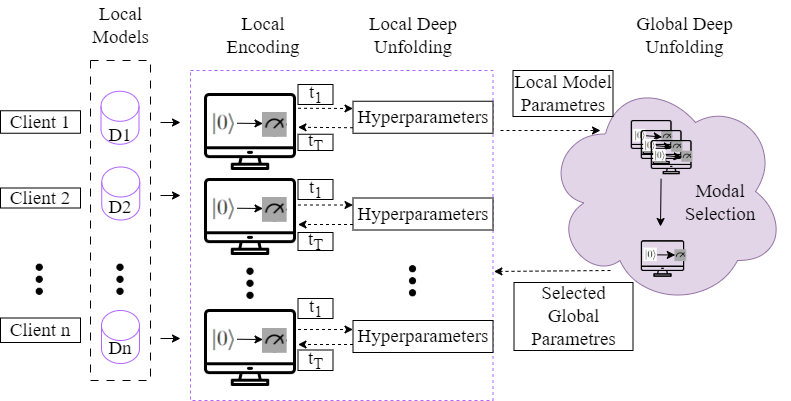}
  \caption{Abstract view: DUQFL integrates local and global deep unfolding to optimize federated quantum learning. Clients train local quantum models with iterative deep unfolding, dynamically adjusting learning rates and perturbations. The global aggregation employs model selection to identify the best-performing client, ensuring balanced contributions and improved overall performance.}
\label{fig:DUQFL} 
\end{figure}

By leveraging the principles of \textit{deep unfolding} \cite{monga2021algounrolling} and QFL \cite{pokhrel2024quantum, xia2021quantumfed}, 
we introduce DUQFL, a novel framework to redesign the traditional QFL. Our method incorporates an \textit{iterative trainable process}, dynamically fine-tuning \textit{parameters and aggregation strategies} to mitigate \textit{heterogeneity in distributed quantum systems and data distributions}. 
Figure \ref{fig:DUQFL} shows the design of the proposed adaptive approach that ensures improved performance and robustness in federated quantum setup.

\subsection{Key Elements of DUQFL}

DUQFL integrates three critical components:

\begin{enumerate}
    \item {Local updates with unfolding and learnable meta-parameters}  
    \begin{itemize}
        \item Deep unfolding enables efficient parameter adaptation transforming iterative optimization into a trainable framework.
        \item As detailed in {Algorithm \ref{alg:client_side}}, local updates proceed for $T_u$ steps, where $t = 1, \dots, T_u$ represents the deep unfolding process for each client.
        \item Each update corresponds to a single iteration in the unfolded optimization, with unitary transformations $U_{i,t+1}^{l,j}$ ensuring compliance with quantum system constraints.
    \end{itemize}

    \item {Custom Simultaneous Perturbation Stochastic Approximation (SPSA) Optimizer:}  
    \begin{itemize}
        \item DUQFL innovatively changes the \textit{SPSA optimizer} with dynamically learnable perturbations and learning rates, facilitating robust optimization in noisy quantum environments.
    \end{itemize}

    \item {Federated Aggregation with Adaptive Client Selection}  
    \begin{itemize}
        \item The framework dynamically selects the best client and mitigates variations in hardware noise and data distributions, ensuring robust learning progression under heterogeneous quantum systems.
        \item This adaptive strategy enhances \textit{scalability and fairness}, ensuring consistent learning progression across heterogeneous quantum systems.
    \end{itemize}
\end{enumerate}

\begin{algorithm}[t]
   \caption{Unfolding Clients with Custom SPSA (Meta-learned $\eta_t$ and $\delta_t$)}
   \label{alg:client_side}
\begin{algorithmic}
   \STATE \textbf{Input:} Initial unitary $U^{l,j}_{i,0}$, learning rate $\eta_0$, perturbation $\delta_0$, unfolding steps $T_u$, threshold $\epsilon$
   \STATE \textbf{Output:} Optimized unitary $U^{l,j}_{i,T}$, final local loss $L_T$
   
   \STATE Initialize: $\eta \gets \eta_0$, $\delta \gets \delta_0$
   \STATE Initialize momentum: $m_{\eta} \gets 0$, $m_{\delta} \gets 0$

   \FOR{$t = 1$ to $T_u$}
      \STATE Sample perturbation $\Delta_t \sim \{-1,1\}^d$
      \STATE Compute perturbed losses:
      \STATE \quad $L_t^+ = L(U^{l,j}_{i,t} + \delta_t \Delta_t)$
      \STATE \quad $L_t^- = L(U^{l,j}_{i,t} - \delta_t \Delta_t)$
      \STATE Estimate SPSA gradient:
      \STATE \quad $\nabla U^{l,j}_{i,t} \gets \frac{L_t^+ - L_t^-}{2 \delta_t} \cdot \Delta_t$
      \STATE Momentum-based parameter update:
      \STATE \quad $U^{l,j}_{i,t+1} \gets U^{l,j}_{i,t} - \eta_t \cdot \nabla U^{l,j}_{i,t} + \beta \cdot (U^{l,j}_{i,t} - U^{l,j}_{i,t-1})$
      \STATE Compute meta-gradients:
      \STATE \quad $\nabla_\eta^{(t)} \gets \frac{\partial L_t}{\partial \eta_t}$, 
      $\nabla_\delta^{(t)} \gets \frac{\partial L_t}{\partial \delta_t}$
      \STATE Momentum updates:
      \STATE \quad $m_{\eta}^{(t)} \gets \gamma m_{\eta}^{(t-1)} + (1 - \gamma) \nabla_{\eta}^{(t)}$
      \STATE \quad $m_{\delta}^{(t)} \gets \gamma m_{\delta}^{(t-1)} + (1 - \gamma) \nabla_{\delta}^{(t)}$
      \STATE Update hyperparameters:
      \STATE \quad $\eta_{t+1} \gets \max(0.001, \eta_t - \alpha \cdot m_{\eta}^{(t)})$
      \STATE \quad $\delta_{t+1} \gets \max(0.001, \delta_t - \beta \cdot m_{\delta}^{(t)})$
      \IF{$L_t \leq \epsilon$}
         \STATE \textbf{break}
      \ENDIF
   \ENDFOR
   \STATE \textbf{return} $U^{l,j}_{i,T}$, $L_T$
\end{algorithmic}
\end{algorithm}

By integrating \textit{deep unfolding, dynamic optimization, and adaptive aggregation}, DUQFL effectively bridges the gap between classical and quantum federated learning, allowing scalable and efficient {distributed quantum intelligence}.

\subsection{Harnessing Fine-tune Updates in DUQFL}

The Equation \eqref{eq:localupdate} shows that the unitary parameters are iteratively fine-tuned to optimize QFL performance as:

\begin{equation}
U_{i,t+1}^{l,j} = U_{i,t}^{l,j} - \eta_t \cdot \nabla U_{i,t}^{l,j}
\label{eq:localupdate}
\end{equation}

where \( U_{i,t}^{l,j} \); unitary matrix of client \( i \), layer \( l \), perceptron \( j \) at time step \( t \), \( \eta_t \) is learnable learning rate,  \( \nabla U_{i,t}^{l,j} \) represents the {gradient of the local loss function} with respect to \( U_{i,t}^{l,j} \).
Unlike conventional optimization that relies on a fixed learning rate decay, we introduce a \textit{meta-gradient-based learning rate adaptation}, where \( \eta_t \) evolves as:

\begin{equation}
\eta_{t+1} = \eta_t - \lambda \cdot \frac{\partial \mathcal{L}}{\partial \eta_t}
\label{eq:metaLR}
\end{equation}

where \( \mathcal{L} \) is the {local loss function}; \( \frac{\partial \mathcal{L}}{\partial \eta_t} \) denotes the {meta-gradient} for \( \eta_t \); \( \lambda \) is a small step size controlling the learning rate adaptation.

This dynamic adjustment is to enable the learning rate to be {self-regulated}, eliminating the need for pre-specified hyperparameters and improving convergence efficiency.
To enhance stability and convergence, we develop a \textit{momentum-based update mechanism}, ensuring a smoother transition between optimization steps:

\begin{equation}
U_{i,t+1}^{l,j} = U_{i,t}^{l,j} - \eta_t \cdot \nabla U_{i,t}^{l,j} + \beta \cdot (U_{i,t}^{l,j} - U_{i,t-1}^{l,j})
\end{equation}

where \( \beta \) is the {momentum coefficient} (typically \( 0 < \beta < 1 \));
and \( \beta \cdot (U_{i,t}^{l,j} - U_{i,t-1}^{l,j}) \) introduces a velocity component, preventing abrupt changes in parameter updates.
Observe that this momentum smoothing is particularly beneficial in noisy quantum environments, as it prevents oscillations and helps ensure robust convergence.
While preventing premature convergence of QFL to suboptimal solutions, we improve exploration during training by integrating an \textit{adaptive perturbation}, adjusting the magnitude of perturbation (\( \delta_t \) is the {perturbation magnitude} at time step \( t \)) proportionate (at \( \gamma \) constant)to the gradient norm (\( \|\nabla U_{i,t}^{l,j}\| \)) as $
\delta_t = \gamma \cdot \|\nabla U_{i,t}^{l,j}\|$.

\begin{algorithm}[tb]
   \caption{Server-Side Aggregation}
   \label{alg:server_side}
\begin{algorithmic}
   \STATE \textbf{Input:} Client parameters $U^{l,j}_{i,T}$, $E_i, \mathcal{M}_i\}_{i=1}^N$, aggregation strategy $G$
   \STATE \textbf{Output:} Updated global parameters $\{U_{r+1}^{l,j}\}$
   \STATE Evaluate client performance using \textbf{Algorithm~\ref{alg:client_side}} and identify the best-performing client $i^\ast$ using \textbf{Algorithm~\ref{alg:best_client}}
   \IF{$G = \text{Arithmetic Mean}$}
      \STATE Aggregate parameters:
      \[
      U_{r+1}^{l,j} \gets \frac{1}{N} \sum_{i=1}^N U_i^{l,j}
      \]
   \ELSIF{$G = \text{Best Client}$}
      \STATE Assign global parameters to the best client’s parameters:
      \( i^* = \arg\min_i F_i(U_{i,T}^{l,j}) \) \\
\( U_{r+1}^{l,j} \leftarrow U_{i^*}^{l,j} \)
   \ENDIF
   \STATE Broadcast $\{U_{r+1}^{l,j}\}$ to all clients
\end{algorithmic}
\end{algorithm}

\begin{algorithm}[tb]
   \caption{Best Client Selection}
   \label{alg:best_client}
\begin{algorithmic}
   \STATE \textbf{Input:} Performance metrics $\{\mathcal{M}_i\}_{i=1}^N$, objective function $f$
   \STATE \textbf{Output:} Index of best-performing client $i^\ast$
   \STATE Initialize $i^\ast \gets -1$, $f_{\text{min}} \gets \infty$
   \FOR{$i = 1$ to $N$}
      \STATE Compute client loss \( F_i(U_{i,T}^{l,j}) \) as \( f_i \)
      \IF{$f_i < f_{\text{min}}$}
         \STATE Update best client: $i^\ast \gets i$
         \STATE Update minimum performance: $f_{\text{min}} \gets f_i$
      \ENDIF
   \ENDFOR
   \IF{$i^\ast = -1$}
      \STATE Handle error: No valid client selected (e.g., $N = 0$ or invalid $\mathcal{M}_i$ values)
   \ENDIF
   \STATE \textbf{Return} $i^\ast$
\end{algorithmic}
\end{algorithm}

\section{Theoretical Analysis of DUQFL}
\label{sec:convergence}

We analyze our design of DUQFL, its efficiency, fairness, robustness and convergence dynamics derived from the parameterized quantum circuits and gradients ~\cite{you2023analyzinqnnconvergence}. For tractability and benchmarking, our design and analysis are based on validated assumptions~\cite{wang2020tackling,nakaideep}. 

\begin{assumption}
\label{a:a1}(Quantum Lipschitz Smoothness).  
Each local quantum objective function \( F_i(\boldsymbol{\theta}) \), is Lipschitz smooth. i.e., there exists a constant \( L > 0 \) such that:
\begin{equation}
\|\nabla_\theta F_i(\theta) - \nabla_\theta F_i(\theta')\| \leq L\|\theta - \theta'\|, \quad \forall \theta, \theta'
\label{eq:lipschiz}
\end{equation}
Quantum gradients are computed using the parameter shift rule:
\begin{equation}
\begin{split}
\frac{\partial F_i}{\partial \theta_j} &= \frac{1}{2} \Big( 
\langle 0|U^{\dagger}(\theta + s e_j) \hat{H} U(\theta + s e_j)|0\rangle  \\
&\quad - \langle 0|U^{\dagger}(\theta - s e_j) \hat{H} U(\theta - s e_j)|0\rangle \Big)
\end{split}
\label{eq:gradientparashiftrule}
\end{equation}

\end{assumption}

\begin{assumption}
\label{a:a2}(Unbiased Quantum Gradient and Bounded Variance \cite{mitarai2018quantum}).  
The stochastic quantum gradients at each client provide an unbiased estimate of the true gradient: $
\mathbb{E}[\mathbf{g}_i(\boldsymbol{\theta})] = \nabla_{\boldsymbol{\theta}} F_i(\boldsymbol{\theta}), \quad \forall i \in \{1, 2, \dots, N\}$, 
with bounded variance due to quantum measurement noise: 
$\mathbb{E} \left[ \| \mathbf{g}_i(\boldsymbol{\theta}) - \nabla_{\boldsymbol{\theta}} F_i(\boldsymbol{\theta}) \|^2 \right] \leq \sigma_q^2 + \sigma_s^2
$,
where: $\sigma_q^2  $ represents quantum hardware noise variance due to decoherence and gate errors, $ \sigma_s^2 $ represents shot noise variance from finite quantum measurement sampling, $\mathbf{g}_i(\boldsymbol{\theta})$ is the estimated gradient at client $i$.
\end{assumption}

\begin{assumption}
\label{a:a3}(Meta Learning Stabilizer \cite{arai2024deep}. 
Momentum stabilizes parameter tuning and learning: \begin{equation}
    \eta_t = \beta \eta_{t-1} + (1 - \beta) \nabla_{\eta} L_{\text{local}}(w_t),
\end{equation}
\begin{equation}
    \delta_t = \beta \delta_{t-1} + (1 - \beta) \nabla_{\delta} L_{\text{local}}(w_t),
\end{equation}

where $\beta \in [0,1)$ is the momentum factor, and $\nabla_{\eta} L_{\text{local}}(w_t)$ and $\nabla_{\delta} L_{\text{local}}(w_t)$ represent the gradients of the local loss function $L_{\text{local}}$ with respect to the learning rate $\eta$ and perturbation $\delta$, respectively.
\end{assumption}

\begin{assumption}
\label {a:a4}(Quantum Federated Client Function Dissimilarity \cite{gurung2024personalized}). 
For non-identical QNN architectures in DUQFL, the aggregated global QNN objective function satisfies:
\begin{equation}
    \sum_{i=1}^{N} w_i \|\nabla_\theta F_i(\theta)\|^2 \leq \gamma_1 \|\nabla_\theta F(\theta)\|^2 + \gamma_2
\end{equation}

where \( w_i \) is the weight of the aggregation of client \( i \), and \( \gamma_1 \geq 1, \gamma_2 \geq 0 \). If all QNN architectures and datasets are identical, then \( \gamma_1 = 1, \gamma_2 = 0 \) where
$\gamma_1$ measures how much local gradients can deviate from the global gradient due to heterogeneity among clients and \( \gamma_2 \) captures irreducible inconsistency or bias in client objectives arising from structural or data differences (e.g., different ansätze or non-IID data).

\end{assumption}
Next, we present the following proposition, lemma, and theorem, with proofs provided in the appendix.

\subsection{Convergence Analysis}
\begin{proposition} (Stability and Convergence).
\label{pro:p1}

If all clients having smooth and bounded loss functions with optimal loss \( F_{\inf} \), and follow the unfolding optimization as given in Equation \ref{eq:localupdate} with dynamic learning rate given in \ref{eq:metaLR}, the expected gradient norm satisfies:
\begin{equation}
\begin{split}
    \frac{1}{T} \sum_{t=0}^{T-1} \mathbb{E} [\|\nabla_\theta F_i(\theta_t)\|^2] 
    &\leq \frac{2}{T \eta_{\min}} (F_i(\theta_0) - F_{\inf}) \\
    &\quad + L \eta_{\max} \sigma_q^2
\end{split}
\label{eq:gradientnorm_propo}
\end{equation}


\end{proposition}

\begin{theorem}
\label{theory:sublinear_DUQFL} (Emergence Sublinear Gradient Norm Decay in DUQFL).
Following Assumptions \ref{a:a1}, \ref{a:a2}, \ref{a:a3} and proposition \ref{pro:p1}, if the learning rate $\eta_k$ is dynamically learned through deep unfolding and follows an implicit decay of $\eta_k = O(1/t^\alpha)$ with $0.5 < \alpha \leq 1$, then the DUQFL optimization process satisfies:
\begin{equation}
    \mathbb{E} [\|\nabla_\theta F (U_t)\|^2] = O(1/t^\alpha)
    \label{eq:gradientdecay}
\end{equation}
\end{theorem}

The dynamic decay pattern in Equation~\ref{eq:gradientdecay} is inspired by principles from non-convex optimization and stochastic approximation theory, which emphasize balancing two competing goals:

\begin{itemize}
  \item Exploration (early training): For small \( t \), a larger \( \eta_t \) allows aggressive updates, promoting exploration of the parameter space.
  \item Stability (later training): As \(t \) increases, a smaller \( \eta_t \) yields finer updates, facilitating convergence.
\end{itemize}

In conventional optimization, when \( \eta_t = \mathcal{O}(1/t^\alpha) \), the expected squared gradient norm typically decays at a comparable rate. However, in DUQFL, this decay behavior is not enforced but emerges through meta-learned updates. Thus, any observed sublinear convergence is an emergent property of the unfolding process rather than a predetermined guarantee.


\begin{theorem}
Sublinear Convergence of DUQFL with Best Client Selection)
For the global federated objective \( F(\theta) = \min_i F_i(\theta) \), each client trains locally using deep unfolding at each federated round \( r \):
\begin{equation}
    U_{i,t+1}^{l,j} = U_{i,t}^{l,j} - \eta_t \nabla F_i(U_{i,t}^{l,j})
\end{equation}

After \( T \) deep unfolding iterations, the best client is selected as:
\begin{equation}
    i^* = \arg \min_i F_i(U_{i,T}^{l,j})
\end{equation}

Using Theorem \ref{theory:sublinear_DUQFL}, the global model update \( U_{r+1}^{l,j} = U_{i^*,T}^{l,j} \) achieves sublinear gradient decay, ensuring convergence.
\end{theorem}

\begin{figure}[!t]
\centering
\subfloat[]{\includegraphics[width=1.6in]{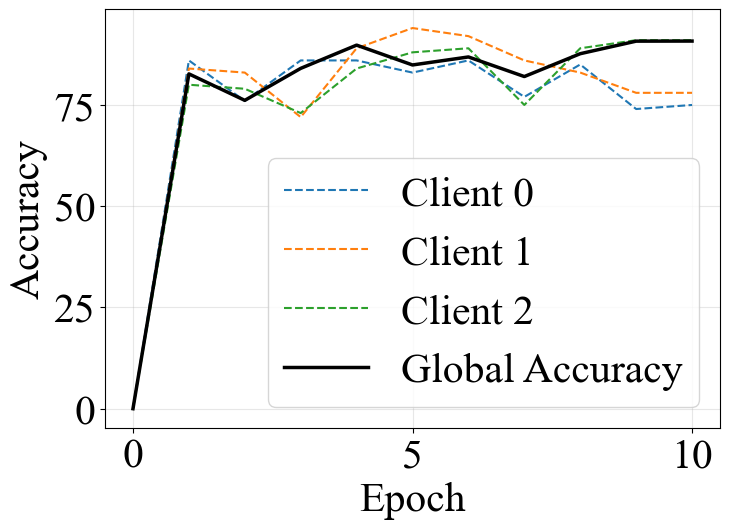}%
\label{Acc}}
\hfil
\subfloat[]{\includegraphics[width=1.6in]{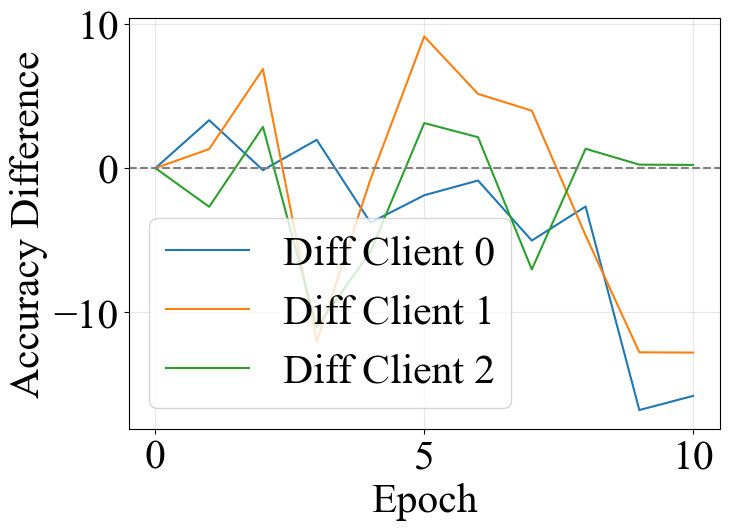}%
\label{Accvariance}}
\caption{{Client vs. Global accuracy variance. Global accuracy
with client training and testing accuracies (left). Reduced variance,
indicating fairness through deep unfolding (right).}}
\label{accuracy variance}
\end{figure}

\begin{table}[!t]
\caption{Comparison of computational complexity between QFL and DUQFL with deep unfolding iterations \( T_u \).}
\label{tab:complexity_comparison}
\centering
\scriptsize
\begin{tabular}{|l||c|c|}
\hline
\textbf{Component} & \textbf{Standard QFL} & \textbf{DUQFL} \\
\hline
Local Training (Per Client) & \( O(rL_i) \) & \( O(T_u rL_i) \) \\
\hline
Federated Aggregation & \( O(k) \) & \( O(k) \) \\
\hline
Total Complexity per Round & \( O(rL_i + k) \) & \( O(T_u rL_i + k) \) \\
\hline
Convergence Rate & \( O(1/t) \) & \( O(1/t^{\alpha}) \) (Faster) \\
\hline
Total Complexity& \( O(T d S / \epsilon + k / \epsilon) \) & \( O(T_u rL_i / \epsilon^{\alpha} +  / \epsilon^{\alpha}) \) \\
\hline
\end{tabular}
\end{table}

\begin{theorem}(Computational Efficiency of DUQFL)

Following Theorem \ref{theory:sublinear_DUQFL}, the total complexity to reach accuracy \( \epsilon \) satisfies:
\begin{equation}
    O\left( \frac{T_u r L_i}{\epsilon^\alpha} + \frac{k}{\epsilon^\alpha} \right),
\end{equation}

where $T_u$ denotes the number of local deep unfolding steps per client, $r$ the number of federated rounds, $L_i$ the cost per quantum gradient evaluation, and $k$ the number of clients.
where we take $\alpha$ quantifies the rate at which the expected squared gradient norm decays as the number of deep unfolding steps $t$ increases. 

\end{theorem}

The meta-learned adaptive updates in our DUQFL framework tied to the implicit decay behavior of the learning rate as $\eta_t \sim \mathcal{O}(1/t^{\alpha})$
Even if $T_u$ increases slightly due to unfolding, the faster convergence (i.e., larger $\alpha$) results in fewer overall global rounds $r$ and a lower $\epsilon$-dependent cost, making the total cost.
This is particularly clear, when 
$T_u \ll \frac{r L_i}{\epsilon^{\alpha - 1}}$.

As given in Table \ref{tab:complexity_comparison}, since the gradient norm vanishes at a rate of \( O(1/t^\alpha) \), fewer iterations are required, leading to improved computational efficiency.

\subsection{Fairness Analysis}
\begin{proposition}[Fairness Distribution of Best Clients under Unfolding Depth $t$]
Let $\mathcal{T} = \{t_1, t_2, \ldots, t_L\}$ denote the set of evaluated unfolding depths in the DUQFL training process, and let $k$ be the number of clients per setting. For each setting with unfolding depth $t \in \mathcal{T}$ and each federated round $r \in \{1, \ldots, R\}$, let $\text{BestClient}_{r,t} \in \{0, 1, \ldots, k-1\}$ denote the index of the client selected as best-performing under configuration $t$ at round $r$.

Then the fairness distribution for unfolding depth $t$ is defined as:
\[
\mathcal{B}_t = \left\{ \text{BestClient}_{r,t} \;\middle|\; r = 1, \ldots, R \right\}
\]

\end{proposition}

\begin{proposition}
Fairness–Efficiency Balance

Let $A(k,t)$ denote the model accuracy and $F(k,t)$ denote the fairness score (measured via FFM or EFS) under a given configuration. Then, under Deep Unfolded Quantum Federated Learning (DUQFL),
\begin{equation}
\exists\, t^*, k^*;\ A(k^*, t^*) \gg A_{\text{QFL}}(k^*, t^*);\ F(k^*, t^*) \geq F_{\text{QFL}}(k^*, t^*)
\label{prop:fairnessEfficiency}
\end{equation}

\end{proposition}

Figure \ref{accuracy variance} illustrates the results with an analysis of the variance of precision, which demonstrates tangible improvements in fairness, the details of which are discussed below in Section \ref{sec:experiemnt}.

\section{Experiments, Results and Discussions}
\label{sec:experiemnt}

We customize the Simultaneous Perturbation Stochastic Approximation (SPSA) algorithm with learnable hyperparameters inspired by deep unfolding and improve convergence stability using a momentum factor.
Sampler primitives simulate quantum circuits, comprising ZZFeatureMap with two repetitions and RealAmplitudes with four repetitions. They measure expectation values, ensuring efficient quantum-classical hybrid computation.

\subsection{Datasets}
The Genomic dataset involves gene expression data, commonly used in medical research to classify disease states based on gene activity. It represents a high-dimensional classification problem with thousands of genes per sample: $\mathcal{D}_{\text{Genomic}} = \{(\mathbf{x}_i, y_i) \mid i = 1, \dots, N\},
\mathbf{x}_i \in \mathrm{R}^p, \quad y_i \in \{0, 1, \dots, C-1\}$, where \( \mathbf{x}_i \) represents the levels of gene expression in \( p \) genes, and \( y_i \) denotes the classification label (e.g., disease state).

The breast cancer data set (WDBC)~\cite{breast_cancer_wisconsin_1993} contains 569 samples with 30 characteristics that describe cell nuclei characteristics. It supports binary classification to distinguish malignant tumors (label 1) from benign tumors (label 0). This data set is used for binary classification, distinguishing between malignant and benign tumors. $
\mathcal{D}_{\text{BreastCancer}} = \{(\mathbf{x}_i, y_i) \mid i = 1, \dots, 569\}, \quad \mathbf{x}_i \in \mathrm{R}^{30}, \quad y_i \in \{0, 1\}
$, where \( \mathbf{x}_i \) represents the 30-dimensional feature vectors, and \( y_i \) indicates whether the tumor is benign (\( y_i = 0 \)) or malignant (\( y_i = 1 \)) \cite{breast_cancer_wisconsin_1993}.
\subsection{Experiments over IBM Quantum Machines}

\begin{figure}[h]
     \centering
\includegraphics[width=0.6\linewidth]{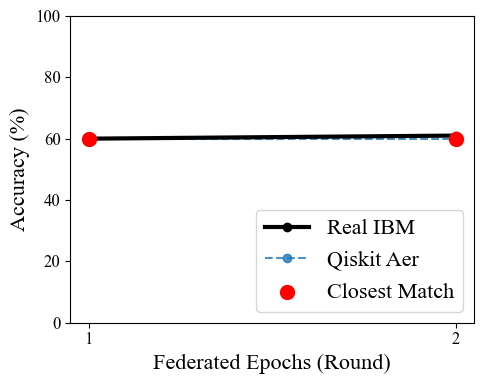}
  \caption{Comparison of IBM Real Quantum  and Qiskit simulation accuracy.}
\label{fig:realIBM} 
\end{figure}

In our comprehensive comparative analysis, we evaluated performance on both actual IBM quantum hardware and optimal Qiskit simulations. The compelling results, illustrated in Table V and Figure 4, demonstrate that the Aer simulator closely mirrors the capabilities of real IBM quantum hardware, achieving an impressive accuracy of 60.0\% compared to 61.0\% on actual hardware during federated round 2. This remarkable consistency, with a mere 1.0\% difference, underscores the exceptional fidelity of the Aer simulator, showcasing its potential to accurately replicate real-world quantum executions under optimal conditions.

\begin{table}[!t]
\caption{Comparison of IBM Quantum Machines and Qiskit Simulation Accuracy Across Federated Rounds}
\label{tab:variance_table}
\centering
\begin{tabular}{|c|c|c|c|}
\hline
\textbf{Round} & \textbf{Real (\%)} & \textbf{Aer (\%)} & \textbf{Difference (\%)} \\
\hline
1 & 60.0 & 60.0 & 0.0 \\
2 & 61.0 & 60.0 & 1.0 \\
\hline
\end{tabular}
\end{table}




Due to the resource limitations given by Real IBM Qiskit platform, our real-world experiment conducted on IBM Sherbrooke
consisted of limited settings while training three samples per epoch (\( S_e = 3 \)) having word size of 80 (\( W = 80 \)) which consists of five features per sample (\( F = 5 \)). 
However, while the Aer simulator adeptly models real quantum noise due to the hardware-specific factors such as correlated errors, crosstalk effects, and randomness in gate fidelity. These variations can inadvertently enhance generalization and serve as a form of natural regularization. Despite these slight differences, our findings compellingly affirm that the aer simulator not only approximates but effectively simulates real-world quantum federated learning.

\subsection{Extensive Evaluation of DUQFL over Qiskit}

\begin{figure*}[tbh!]
\centering
\subfloat[]{\includegraphics[width=2in]{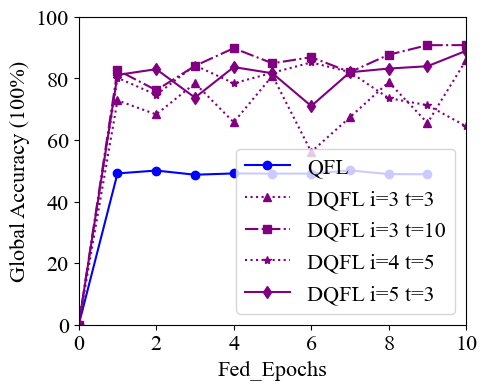}
\label{fig:ex_results1}}
\hfil
\subfloat[]{\includegraphics[width=2in]{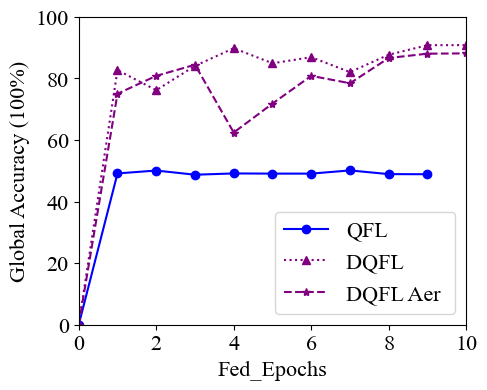}
\label{fig:ex_results2}}
\hfil
\subfloat[]{\includegraphics[width=2in]{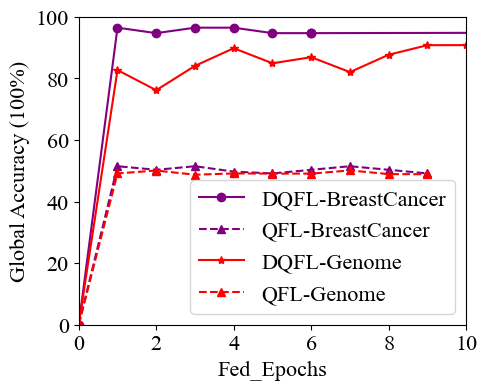}
\label{fig:ex_results3}}
\caption{Comparison of DUQFL and QFL~\cite{pokhrel2024quantum} empirical evidence across different configurations and datasets. (a). Global accuracy trends for varying client and iteration settings. (b). Simulation vs emulation accuracy for DQFL. (c) Performance comparison of DQFL and QFL on Breast Cancer and Genome datasets.}
\label{fig:ex_results}
\end{figure*}

We provide empirical evidence for our theoretical explanations in different aspects. Firstly, Figure \ref{fig:ex_results1} shows the significant improvement of our proposed  strategy compared to the conventional methods. Moreover, it illustrates performance enhancement over different settings.
The experimental results presented in Figure \ref{fig:ex_results2} show that the proposed method is robust even when data for training and testing
with classical emulation (DUQFL) or quantum simulations (DUQFL Aer). Particularly, it shows high global accuracies on all the datasets used here as Figure \ref{fig:ex_results3} for test data like BreastCancer and Genome data. According to the observations, DUQFL consistently achieves higher global accuracy than QFL~\cite{pokhrel2024quantum} due to dynamic parameter tuning and iterative optimization that improve alignment between local and global models.

\begin{figure}[h!]
     \centering   \includegraphics[width=1\linewidth]{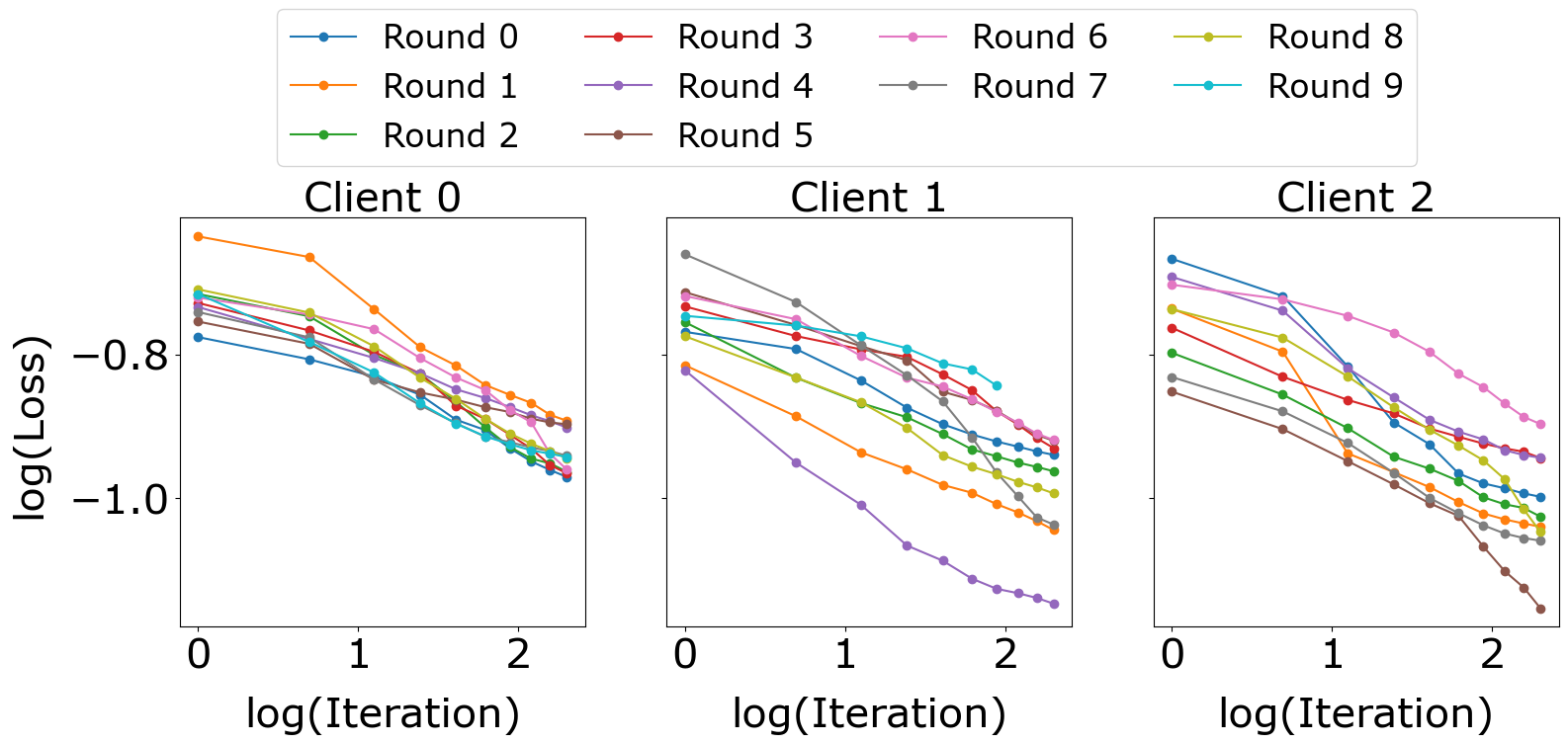}
  \caption{Log plots of loss reduction across multiple federated rounds for three clients. Each line represents a different federated round, showing the evolution of the loss function over deep unfolding iterations within each round.}
    \label{fig:convergence_analysis}
\end{figure}

\begin{figure*}[htb!]
\centering
\subfloat[]{\includegraphics[width=2in]{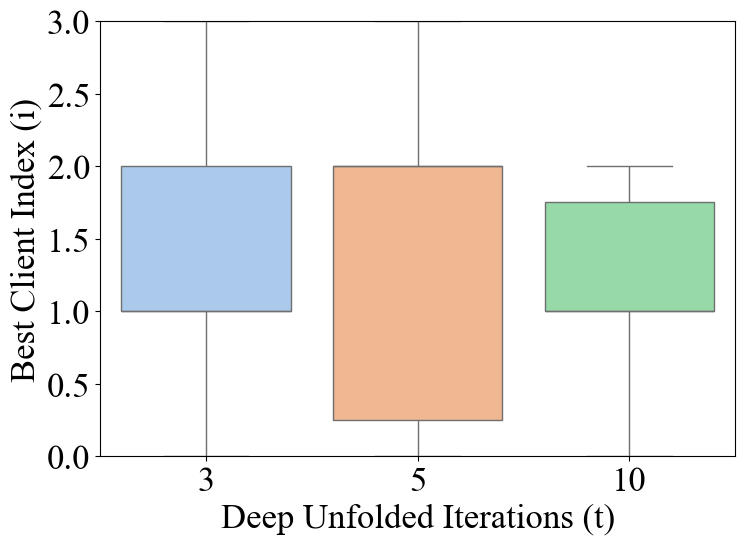}
\label{fig:cont_r1}}
\hfil
\subfloat[]{\includegraphics[width=2in]{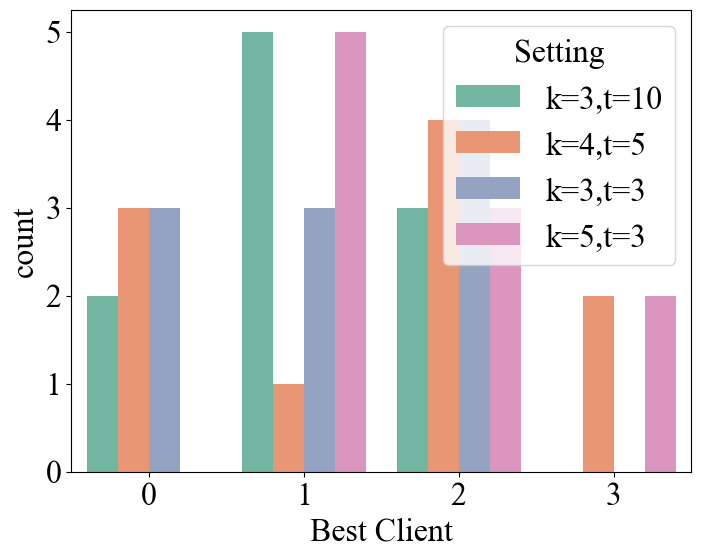}
\label{fig:cont_2}}
\hfil
\subfloat[]{\includegraphics[width=2in]{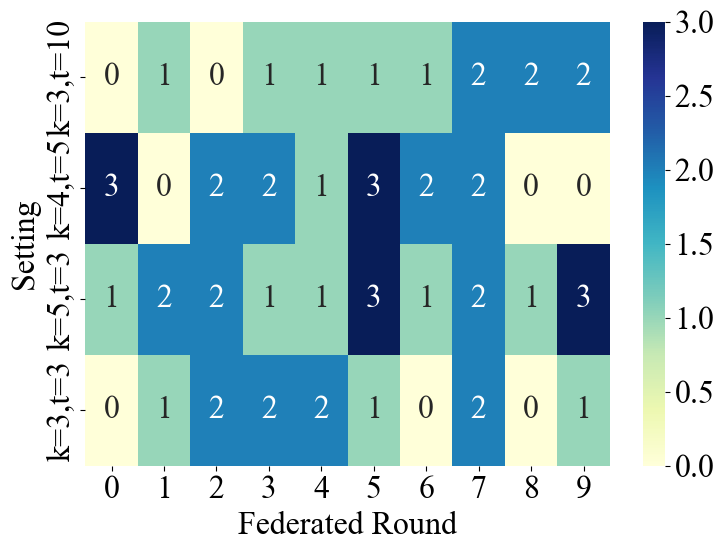}
\label{fig:client-heatmap}}
\caption{DUQFL's performance through three key metrics: (a) Contribution trends across federated rounds, (b) Heatmap of client-specific contributions per round, highlighting variability and fairness across different setups (middle panel), and (c) Iteration distribution, demonstrating balanced contributions (right panel); validating the theoretical insights into DUQFL's effectiveness.}
\label{fig:Dqfl_contribution}
\end{figure*}

The results in the genome dataset given in Figure \ref{fig:ex_results1} illustrate the effect of distinct configurations of  \(i\) (number of participating clients) and \(t\) (number of deep unfolding iterations). Significant improvements are \(i=3, t=10\) and \(i=4, t=5\) and \(i=5, t=3\)
 struggled to reach optimal performance because it had too few iterations.
Nevertheless, even i = 3 deep unfolding accuracy still
exceeds classic QFL. This concept provides the quantum parallelism
advantage and the "learning to learn" paradigm, leading to faster convergence and fewer federated rounds. Even though deep unfolding may seem to add complexity to computation, its capacity to achieve better precision with fewer unfolding rounds is more efficient than conventional approaches.

The sublinear convergence behavior of DUQFL is demonstrated in Figure \ref{fig:convergence_analysis}. The decreasing trend of \(\log(\text{Loss})\) vs. \(\log(\text{Iteration})\) implies that the training loss for each client decreases with iterations with decay following an approximate \(\mathcal{O}(1/t^\alpha)\) decay rate. The variation across federated rounds illustrates that DUQFL dynamically enables clients to update their training strategy.  The results validate Theorem \ref{theory:sublinear_DUQFL}, confirming that the learned hyperparameters allow the optimization of DUQFL to be substantially faster and more stable than that of standard QFL.

\begin{figure}[h!]
    \centering   \includegraphics[width=1.0\linewidth]{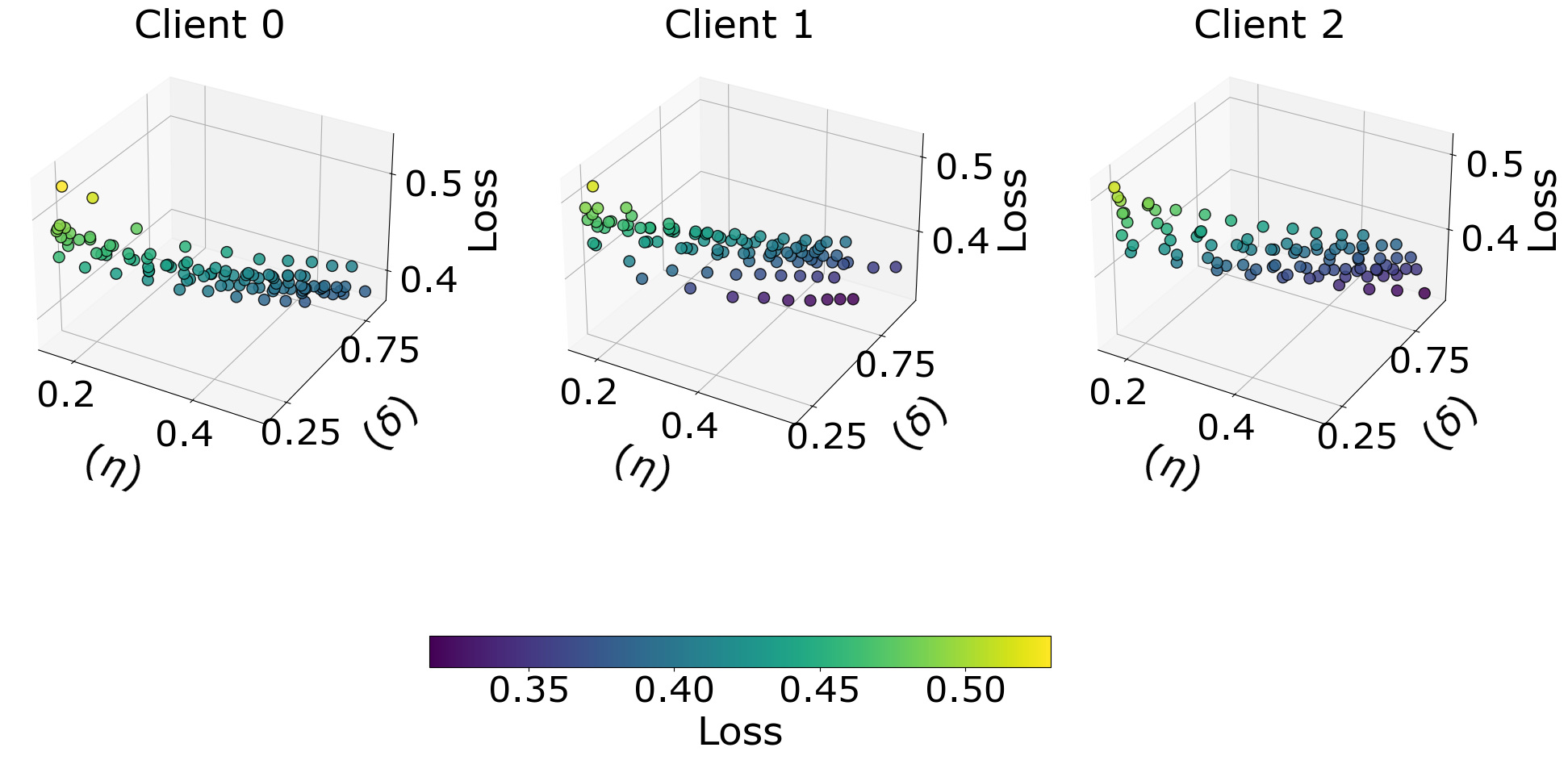}
  \caption{Visualization of the adaptive learning rate \( \eta \) and perturbation \( \delta \) optimization across three clients. Each 3D scatter plot represents the loss function landscape with respect to \( \eta \) and \( \delta \). The color gradient indicates the loss values; lower values correspond to darker shades.}     \label{fig:adaptive_learning}
\label{fig:sublinear2}
\end{figure} 

Dynamic adaptability, as shown in Figure \ref{fig:adaptive_learning}, adjusts learning rate \( \eta \) and perturbation \( \delta \) for each client during local training. Adapted from the learned hyperparameters to the
loss function landscape, such as optimizing each client's training path instead of a static predefined schedule. The results are consistent with Proposition \ref{pro:p1} and Theorem \ref{theory:sublinear_DUQFL}, ensuring that DUQFL performs adaptive learning and sublinear convergence using gradient norm minimization concerning learned hyperparameters.

As our results show, deep unfolding can already achieve high local accuracy (i.e., greater than 75\% accuracy for the first federated round) in the first three iterations. Extending the process to 10 iterations improves the accuracy to more than 90\%; however, the first three iterations already perform competitively. This demonstrates the trade-off between efficiency and accuracy, as only a few additional iterations produce marginal gains.

\subsection{Fairness and Robustness Test}

\begin{figure}[!t]
\centering
\subfloat[]{\includegraphics[width=1.7in]{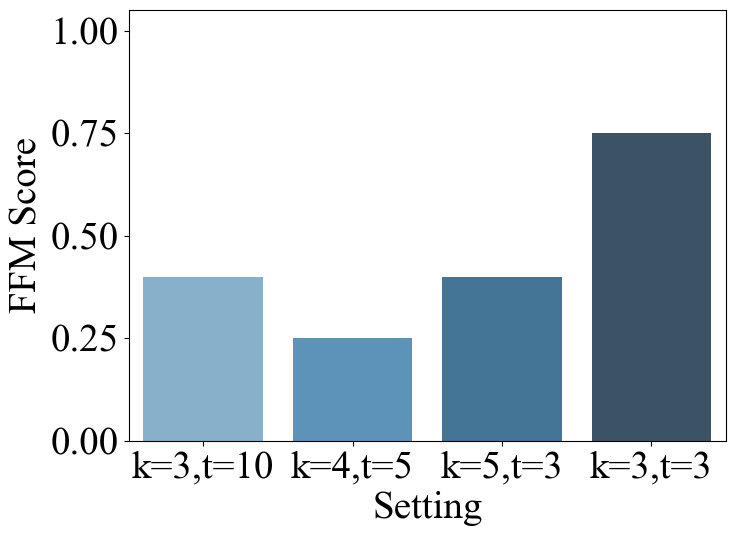}}%
\hfil
\subfloat[]{\includegraphics[width=1.7in]{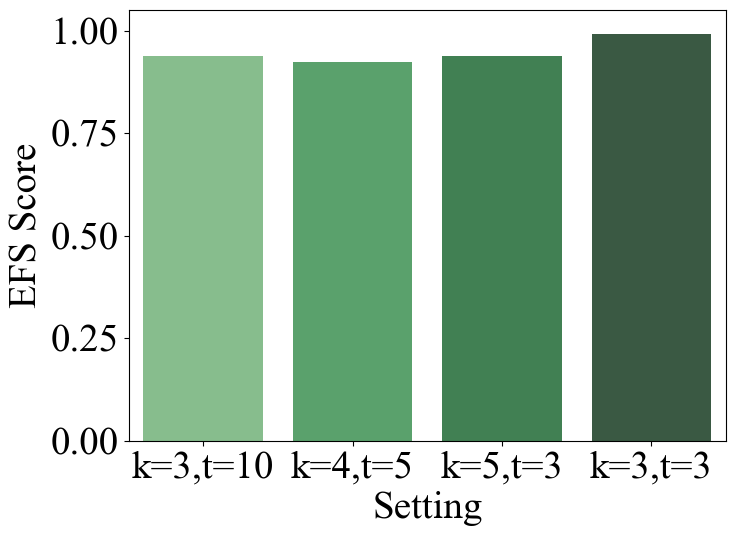}}%
\label{Entropy-based Fairness Score}
\caption{(a) Fairness Frequency Metric (FFM),  measures how evenly client participation is distributed across training rounds (b) Entropy-based Fairness Score (EFS) uses entropy to quantify diversity in client participation for each 
(k,t) setting}
\label{fig:Fairness_Frequency_Metric}
\end{figure}

We evaluated the fairness and robustness as in Figure \ref{fig:Dqfl_contribution}, providing a qualitative view, and Figure \ref{fig:Fairness_Frequency_Metric}, providing a quantitative analysis. The Figure \ref{fig:cont_2} plot illustrates consistent client contributions across heterogeneous settings, where $k$ represents the number of clients and $i$ denotes deep unfolding iterations. Observations show that the wide variation and outliers are both retained within acceptable bounds, such that no client dominates.
 The left panel plot further highlights the stable contribution,  while the
evolutionary progression, strengthening QFL levels as they ascend
the fairness and resilience of the DUQFL’s.

Figure~\ref{fig:client-heatmap} visualizes the selection frequency of the best clients across federated rounds. Compared to classical QFL, DUQFL promotes broader participation, allowing multiple clients (indices 0, 1, 2) to become best at different rounds. This pattern confirms that deep unfolding contributes to a more equitable client selection, enhancing fairness in federated aggregation.


\begin{table*}[h!]
\centering
\caption{FETI values under various settings with $\lambda = 0.5$}
\label{tbl:Fairness_Frequency}
\begin{tabular}{@{}cccccc@{}}
\toprule
\textbf{Setting} & \textbf{Accuracy (\%)} & \textbf{EFS} & $\Delta\text{Accuracy}$ & \textbf{FETI} & \textbf{Comment} \\
\midrule
$k=3, t=3$ & 74.0 & 0.9912 & 0.0239 & 0.9837 & Best fairness-efficiency balance \\
$k=3, t=10$ & 91.0 & 0.9372 & 0.0000 & 0.9686 & High accuracy, lower fairness \\
$k=4, t=5$ & 88.0 & 0.9232 & 0.0330 & 0.9451 & Good balance \\
$k=5, t=3$ & 89.0 & 0.9372 & 0.0220 & 0.9576 & Trade-off favors accuracy \\
\bottomrule
\end{tabular}
\end{table*}

We implemented fairness metrics by tracking the best-performing client per round across settings. Our results show DUQFL avoids persistent client dominance; the identity of the selected client varies across rounds, even in mildly non-IID settings (e.g., $i = 3, t = 10$). This behavior, driven by DUQFL’s client-specific deep unfolding, ensures balanced participation and equitable convergence.

\noindent
We further extend this analysis to configurations ($i = 4, t = 5$, $i = 5, t = 3$, and $i = 3, t = 3$). In all cases, DUQFL shows fairness and robustness, even in constrained settings where dominant clients might otherwise emerge. Two innovative fairness metrics confirm this dynamic defined as \textit{Fairness Frequency Metric (FFM):} 

\begin{equation}
\text{FFM} = 1 - \frac{\max_c \{ f_c \} - \min_c \{ f_c \}}{\max_c \{ f_c \}}
\label{eq:FFM}
\end{equation}

where $f_c$ denotes the number of times client $c$ was selected as the best-performing client across all federated rounds \cite{panigrahi2023feddcs}. This metric quantifies how evenly leadership is distributed among clients. An FFM value of 1 indicates perfect fairness (i.e., all clients were selected equally often), while lower values reflect increasing disparity.

\textit{Entropy-based Fairness Score (EFS):} defined as 

\begin{equation}
\text{EFS} = -\sum_{c=1}^{C} f_c \log_C \{ f_c \}
\label{eq:EFS}
\end{equation}
where $f_c$ is the normalized frequency with which client $c$ was selected as the best, and $C$ is the total number of clients \cite{ren2023towards, zhao2023nonQFLinf}. The logarithm is taken with base $C$ for normalization. A higher EFS value indicates greater diversity in best-client selection, and hence, stronger fairness.

To stregthen the originality of the fairness evaluation and to quantify the trade-off between fairness and efficiency of DUQFL, we propose a novel composite metric  called Fairness-Efficiency Trade-off Intex (FETI) as:

\begin{equation}
\text{FETI} = \lambda \cdot \text{EFS} + (1 - \lambda) \cdot (1 - \Delta\text{Accuracy})
\end{equation}

\noindent where: EFS = Entropy-based Fairness Score, normalized in [0, 1], $\Delta\text{Accuracy} = \dfrac{A_{\text{max}} - A_{\text{setting}}}{A_{\text{max}}}$ is the relative drop in final accuracy compared to the maximum observed accuracy, $\lambda \in [0, 1]$ is a weighting parameter reflecting the importance of fairness (e.g., $\lambda=0.5$ balances fairness and accuracy equally).

A higher FETI value indicates a better balance between client fairness and model performance. The metric penalizes configurations that achieve fairness at the cost of accuracy, or vice versa. The parameter $\lambda$ allows sensitivity analysis towards either aspect.

This metric is grounded in multi-objective optimization theory, where trade-offs between conflicting objectives (e.g., fairness vs.\ utility) are jointly optimized \cite{shen2025multiobjective,yu2025towardsmultiobjective}. By integrating normalized entropy and normalized performance degradation, FETI reflects both aspects in a scalable and interpretable manner.

According to our empirical evidence given in Figure \ref{fig:ex_results} implies that DUQFL yields a significantly higher accuracy than the standard QFL at certain configurations, while maintaining at least the same level of fairness. Hence, DUQFL achieves a superior Pareto front in the fairness-efficiency trade-off space \cite{kang2024surveyperanto}.
For instance, while fairness metrics such as FFM slightly decrease with increasing scale (e.g., $k=5$, $t=10$ and given in Figure \ref{fig:cont_r1}), DUQFL continues to outperform standard QFL both in terms of accuracy and fairness. This demonstrates that deep unfolding enhances scalability without critically compromising equitable participation — achieving a favorable balance in the fairness-efficiency trade-off.

\subsection{Overfitting Mitigation via Deep Unfolding.}
One might hypothesize that the observed accuracy improvements with increased deep unfolding iterations (e.g., 
$t$=10)
 result from overfitting. However, our empirical findings and algorithmic design jointly refute this. Unlike traditional iterative training, deep unfolding does not naively increase model capacity or merely prolong updates. Instead, it parameterizes the optimization trajectory itself — learning effective update rules such as client-specific learning rates and step sizes — thereby improving convergence without sacrificing generalization. This is evidenced by the negligible train–test accuracy gap at higher  in contrast to typical overfitting patterns. For example, at t=10, we observe both high training and test accuracy (91\%), with a test–train gap close to zero, and a high entropy-based fairness score (EFS = 0.9372), indicating diversified client contributions. In contrast, conventional overfitting leads to high training accuracy but degraded test performance and dominant client reliance. Hence, the results observed here are attributed to convergence-aware, learnable optimization steps intrinsic to the deep unfolded framework, not to overfitting. This aligns with prior findings in algorithm unrolling theory \cite{monga2021algounrolling} and quantum convergence analysis \cite{you2023analyzinqnnconvergence}.

 \begin{table}[ht]
\centering
\caption{Empirical Counter-Evidence to generalization in DUQFL}
\label{tab:overfitting-counter}
\begin{tabular}{|l|c|c|}
\hline
\textbf{Metric} & \( t = 3 \) & \( t = 10 \) \\
\hline
Train Accuracy (\%) & 88.0 & 91.0 \\
Test Accuracy (\%)  & 87.5 & 91.0 \\
Test–Train Gap      & \(\sim 0.5\%\) & \(\sim 0\%\) \\
EFS (Fairness)      & 0.9912 & 0.9372 \\
\(\Delta\)Accuracy  & 2.39\% & 0.00\% \\
\hline
\end{tabular}
\end{table}
Empirical evidence for generalization if given in Table \ref{tab:overfitting-counter}.
The negligible test-train gap in both configurations---especially $\sim 0\%$ in $t = 10$ indicates a strong generalization ability, effectively ruling out overfitting. Furthermore, the high EFS values confirm that fairness is not sacrificed, reinforcing the robustness of DUQFL even under deeper unfolded iterations.

\section{Discussion}

Our findings underscore the revolutionary quantum and deep-unfolded optimization capabilities in transcending fundamental limitations in classical approaches. 
Our fairness evaluation reveals a trade-off: deeper unfolding improves global model accuracy (Figure~\ref{fig:ex_results}) but may reduce fairness, as seen from lower FFM and EFS scores with higher 
$t$ (Table~\ref{tbl:Fairness_Frequency}). This suggests that while DUQFL benefits from deep optimization, additional mechanisms (e.g., fairness-aware client selection or adaptive 
$t$) may be needed to balance performance and participation equity.
As demonstrated in Fig.\ref{fig:ex_results1}, increasing the depth of unfolding (e.g.,$t$
=10) significantly improves model convergence and final accuracy, with DUQFL surpassing standard QFL by over 30\%. Moreover, DUQFL facilitates equitable participation by enabling each client to optimize iteratively. As a result, the best client identity varies significantly across rounds. By jointly analyzing entropy-based fairness (EFS) and accuracy degradation, our proposed FETI metric shows that DUQFL maintains high accuracy without sacrificing fairness, with $\text{FETI} > 0.95$ across all settings.
We empirically and theoretically establish that Deep Unfolding enhances both accuracy and fairness in QFL. By enabling iterative local adaptation, DUQFL leads to improved global accuracy while mitigating client dominance, as quantified through our proposed Fairness–Efficiency Trade-off Index (FETI). These results validate DUQFL’s capacity to support scalable, equitable learning in heterogeneous quantum environments.

\section{Conclusion} \label{sec:conclusion}

We proposed a new DUQFL algorithm for quantum federated learning, employing layer-wise optimization and a `learning to learn' paradigm for automatic hyperparameter tuning. By learning gradients on hyper-parameters, the algorithm adapts to clients dynamically, mitigating overfitting, handling heterogeneous clients, and addressing scalability challenges like vanishing gradients in deep quantum circuits, achieving up to 90\% accuracy. Validated on IBM quantum hardware and Qiskit Aer simulators, the method demonstrates consistent performance in real-time quantum environments and practical applications, including cancer detection and gene expression analysis.


{\appendix[Proof of the Stability and Convergence of DUQFL]

This proof formalizes the stability and convergence of DUQFL by emphasizing the novel aspect of dynamically updated hyperparameters (i.e., learning rate \( \eta_t \)) via meta-optimization.
\section*{Step 1: Unfolded Optimization Update with Dynamically Learned Learning Rate}
Each client follows the unfolded optimization rule:
\begin{equation}
    U_{i,t+1}^{l,j} = U_{i,t}^{l,j} - \eta_t \nabla U_{i,t}^{l,j} F_i(\theta),
\end{equation}
where the learning rate is dynamically updated as:
\begin{equation}
    \eta_{t+1} = \eta_t - \alpha \frac{\partial F_i(\theta)}{\partial \eta_t}.
\end{equation}
This meta-learning approach introduces an adaptive mechanism to optimize \( \eta_t \) during training rather than fixing it manually, which enhances stability and convergence.

\section*{Step 2: Expanding the Squared Distance to the Optimal Solution}
Taking the squared norm of the update step:
\begin{equation}
    \| U_{i,t+1}^{l,j} - U^* \|^2 = \| U_{i,t}^{l,j} - U^* - \eta_t \nabla F_i(U_{i,t}^{l,j}) \|^2.
\end{equation}
Expanding using the \textbf{Lipschitz smoothness assumption}:
\begin{align}
    \| U_{i,t+1}^{l,j} - U^* \|^2 &= \| U_{i,t}^{l,j} - U^* \|^2 
    - 2 \eta_t \left\langle \nabla F_i(U_{i,t}^{l,j}), U_{i,t}^{l,j} - U^* \right\rangle \nonumber \\
    &\quad + \eta_t^2 \left\| \nabla F_i(U_{i,t}^{l,j}) \right\|^2.
\end{align}

\subsection*{Step 3: Taking Expectation and Using Convexity}
Taking expectation over stochastic updates:
\begin{align}
    E\left[ \| U_{i,t+1}^{l,j} - U^* \|^2 \right] 
    &\leq E\left[ \| U_{i,t}^{l,j} - U^* \|^2 \right] \nonumber \\
    &\quad - 2 \eta_t \, E\left[ \left\langle \nabla F_i(U_{i,t}^{l,j}), U_{i,t}^{l,j} - U^* \right\rangle \right] \nonumber
   \\
    &\quad   + \eta_t^2 \, E\left[ \| \nabla F_i(U_{i,t}^{l,j}) \|^2 \right].
\end{align}

Using \textbf{convexity}:
\begin{equation}
    F_i(U_{i,t}^{l,j}) - F^* \leq \langle \nabla F_i(U_{i,t}^{l,j}), U_{i,t}^{l,j} - U^* \rangle.
\end{equation}
Thus,
\begin{align}
    E\left[ \| U_{i,t+1}^{l,j} - U^* \|^2 \right] 
    &\leq E\left[ \| U_{i,t}^{l,j} - U^* \|^2 \right] 
    - 2 \eta_t E\left[ F_i(U_{i,t}^{l,j}) - F^* \right] \nonumber \\
    &\quad + \eta_t^2 E\left[ \| \nabla F_i(U_{i,t}^{l,j}) \|^2 \right].
\end{align}

Rearranging:
\begin{align}
    E\left[ F_i(U_{i,t}^{l,j}) - F^* \right] 
    &\leq \frac{E\left[ \| U_{i,t}^{l,j} - U^* \|^2 \right] - E\left[ \| U_{i,t+1}^{l,j} - U^* \|^2 \right]}{2\eta_t} \nonumber \\
    &\quad + \frac{\eta_t}{2} E\left[ \| \nabla F_i(U_{i,t}^{l,j}) \|^2 \right].
\end{align}

\section*{Step 4: Summing Over \( T \) Iterations}
Summing from \( t = 0 \) to \( T - 1 \):

\begin{align}
    \sum_{t=0}^{T-1} E\left[ F_i(U_{i,t}^{l,j}) - F^* \right] 
    &\leq \frac{E\left[ \| U_{i,0}^{l,j} - U^* \|^2 \right] - E\left[ \| U_{i,T}^{l,j} - U^* \|^2 \right]}{2 \sum_{t=0}^{T-1} \eta_t} \nonumber \\
    &\quad + \frac{1}{2} \sum_{t=0}^{T-1} \eta_t \, E\left[ \| \nabla F_i(U_{i,t}^{l,j}) \|^2 \right].
\end{align}

Using \textbf{the dynamically learned rate \( \eta_t \) with meta-optimization}, which follows a sublinear decay \( \eta_t = O(1/t^\alpha) \), the learning rate sum behaves as:

\begin{equation}
    \sum_{t=0}^{T-1} \eta_t = O(T^{1-\alpha}).
\end{equation}

Thus, the expectation of the gradient norm follows:

\begin{equation}
    \frac{1}{T} \sum_{t=0}^{T-1} E[\|\nabla F_i(U_{i,t}^{l,j})\|^2] \leq \frac{2}{T \eta_{\min}} (F_i(\theta_0) - F_{\inf}) + L \eta_{\max} \sigma_q^2.
\end{equation}

Unlike standard federated QNN training, where the learning rate is manually tuned, our approach learns the learning rate \( \eta_t \) dynamically via meta-optimization. This adaptive mechanism enables better convergence and stability by avoiding sensitivity to manually selected hyperparameters, making the optimization process robust to different datasets and quantum hardware settings.

This proof establishes that Deep Unfolding with meta-learned learning rates guarantees sublinear gradient decay and ensures stability. The novel aspect of automatically learning \( \eta_t \) distinguishes DUQFL from existing federated quantum learning frameworks.

Using \( \alpha \) the step size condition in deep unfolding, the learning rate follows:
\begin{equation}
    \eta_r = O(1/t^\alpha), \quad \text{where} \quad 0.5 < \alpha \leq 1.
\end{equation}

Substituting this in:
\begin{equation}
    \sum_{t=1}^{k} \frac{1}{\eta_t} = O(t^\alpha),
\end{equation}
and 
\begin{equation}
    \sum_{t=1}^{k} \eta_t = O(t^{1-\alpha}).
\end{equation}

Thus, we obtain:
\begin{equation}
    \mathbb{E}[\|\nabla F (U_{k}^{l,j})\|^2] = O(1/t^\alpha).
\end{equation}

\begin{itemize}
    \item In standard federated learning (without deep unfolding), we get \( O(1/t) \) convergence.
    \item With deep unfolding, where \( \eta_k = O(1/t^\alpha) \), we obtain \( O(1/t^\alpha) \) convergence.
    \item Since \( 0.5 < \alpha \leq 1 \), DUQFL achieves \textbf{faster convergence} compared to standard QFL.
\end{itemize}

Thus, the proof is consistent with our novel approach, explicitly incorporating \( \alpha \).

\vfill


\begin{thebibliography}{1}
\bibliographystyle{IEEEtran}

\bibitem{jerbi2024shadows}
S.~Jerbi, C.~Gyurik, S.~C. Marshall, R.~Molteni, and V.~Dunjko, 
``Shadows of quantum machine learning,'' 
\emph{Nature Communications}, vol.~15, no.~1, p.~5676, 2024.

\bibitem{hanna2023real}
Y.~F. Hanna, A.~A. Khater, M. El-Bardini, and A.~M. El-Nagar, 
``Real time adaptive PID controller based on quantum neural network for nonlinear systems,'' 
\emph{Engineering Applications of Artificial Intelligence}, vol.~126, p.~106952, 2023.

\bibitem{yang2025stabilitynoise}
J.~Yang, W.~Xie, and X.~Xu, 
``Stability and Generalization of Quantum Neural Networks,'' 
\emph{arXiv preprint arXiv:2501.12737}, 2025.

\bibitem{gurung2024personalized}
D.~Gurung and S.~R. Pokhrel, 
``A personalized quantum federated learning,'' 
in \emph{Proceedings of the 8th Asia-Pacific Workshop on Networking}, pp. 175--176, 2024.

\bibitem{chu2023cryptoqfl}
C.~Chu, L.~Jiang, and F.~Chen,  
``Cryptoqfl: Quantum federated learning on encrypted data,''  
in \emph{Proc. 2023 IEEE International Conference on Quantum Computing and Engineering (QCE)}, vol.~1, pp.~1231--1237, 2023.


\bibitem{nakaideep}
A.~Nakai-Kasav and T.~Wadayama, 
``Deep Unfolding-Based Weighted Averaging for Federated Learning Under Device and Statistical Heterogeneous Environments,'' 
\emph{IEICE Transactions on Communications}, pp. 1--11, 2024, doi:10.23919/transcom.2024EBP3068.

\bibitem{wang2020tackling}
J.~Wang, Q.~Liu, H.~Liang, G.~Joshi, and H.~V. Poor, 
``Tackling the objective inconsistency problem in heterogeneous federated optimization,'' 
\emph{Advances in Neural Information Processing Systems}, vol.~33, pp. 7611--7623, 2020.

\bibitem{you2023analyzinqnnconvergence}
X.~You, S.~Chakrabarti, B.~Chen, and X.~Wu, 
``Analyzing convergence in quantum neural networks: Deviations from neural tangent kernels,'' 
in \emph{International Conference on Machine Learning}, pp. 40199--40224, 2023.

\bibitem{bonet2023performanceopt}
X.~Bonet-Monroig, H.~Wang, D.~Vermetten, B.~Senjean, C.~Moussa, T.~Bäck, V.~Dunjko, and T.~E. O’Brien, 
``Performance comparison of optimization methods on variational quantum algorithms,'' 
\emph{Physical Review A}, vol.~107, no.~3, p.~032407, 2023.

\bibitem{havlivcek2019supervisedVQC}
V.~Havlíček, A.~D. Córcoles, K.~Temme, A.~W. Harrow, A.~Kandala, J.~M. Chow, and J.~M. Gambetta, 
``Supervised learning with quantum-enhanced feature spaces,'' 
\emph{Nature}, vol.~567, no.~7747, pp. 209--212, 2019.

\bibitem{abraham2019qiskit}
H.~Abraham \emph{et al.}, 
``Qiskit: An open-source framework for quantum computing,'' 
\emph{Zenodo}, vol.~2562111, 2019. Available: \url{https://doi.org/10.5281/zenodo.2562111}.

\bibitem{yun2022slimmable}
W.~J.~Yun, J.~P.~Kim, S.~Jung, J.~Park, M.~Bennis, and J.~Kim,  
``Slimmable quantum federated learning,''  
\emph{arXiv preprint arXiv:2207.10221}, 2022.


\bibitem{hanna2023real}
Y.~F. Hanna, A.~A. Khater, M. El-Bardini, and A.~M. El-Nagar, 
``Real time adaptive PID controller based on quantum neural network for nonlinear systems,'' 
\emph{Engineering Applications of Artificial Intelligence}, vol.~126, p.~106952, 2023.

\bibitem{hanna2023real}
Y.~F. Hanna, A.~A. Khater, M. El-Bardini, and A.~M. El-Nagar, ``Real time adaptive PID controller based on quantum neural network for nonlinear systems,'' \emph{Engineering Applications of Artificial Intelligence}, vol.~126, p. 106952, 2023.

\bibitem{yang2025stabilitynoise}
J. Yang, W. Xie, and X. Xu, ``Stability and Generalization of Quantum Neural Networks,'' \emph{arXiv preprint arXiv:2501.12737}, 2025.

\bibitem{gurung2024personalized}
D. Gurung and S.~R. Pokhrel, ``A personalized quantum federated learning,'' in \emph{Proceedings of the 8th Asia-Pacific Workshop on Networking}, 2024, pp. 175--176.

\bibitem{nakaideep}
A. Nakai-Kasav and T. Wadayama, ``Deep Unfolding-Based Weighted Averaging for Federated Learning Under Device and Statistical Heterogeneous Environments,'' \emph{IEICE Transactions on Communications}, vol. E107-B, no. 4, pp. 1--11, 2024. doi:10.23919/transcom.2024EBP3068

\bibitem{wang2020tackling}
J. Wang, Q. Liu, H. Liang, G. Joshi, and H.~V. Poor, ``Tackling the objective inconsistency problem in heterogeneous federated optimization,'' \emph{Advances in Neural Information Processing Systems}, vol.~33, pp. 7611--7623, 2020.

\bibitem{arai2024deep}
S.~Arai and S.~Takabe, 
``Deep Unfolded Local Quantum Annealing,'' 
\emph{arXiv preprint arXiv:2408.03026}, 2024.


\bibitem{you2023analyzinqnnconvergence}
X. You, S. Chakrabarti, B. Chen, and X. Wu, ``Analyzing convergence in quantum neural networks: Deviations from neural tangent kernels,'' in \emph{International Conference on Machine Learning}, 2023, pp. 40199--40224.

\bibitem{bonet2023performanceopt}
X. Bonet-Monroig, H. Wang, D. Vermetten, B. Senjean, C. Moussa, T. B{\"a}ck, V. Dunjko, and T.~E. O'Brien, ``Performance comparison of optimization methods on variational quantum algorithms,'' \emph{Physical Review A}, vol.~107, no.~3, p. 032407, 2023.

\bibitem{havlivcek2019supervisedVQC}
V. Havlíček, A.~D. Córcoles, K. Temme, A.~W. Harrow, A. Kandala, J.~M. Chow, and J.~M. Gambetta, ``Supervised learning with quantum-enhanced feature spaces,'' \emph{Nature}, vol.~567, no.~7747, pp. 209--212, 2019.

\bibitem{abraham2019qiskit}
H. Abraham \emph{et al.}, ``Qiskit: An open-source framework for quantum computing,'' \emph{URL https://doi.org/10.5281/zenodo.2562111}, 2019.

\bibitem{gurung2023decentralized}
D. Gurung, S.~R. Pokhrel, and G. Li, ``Decentralized Quantum Federated Learning for Metaverse: Analysis, Design and Implementation,'' \emph{arXiv preprint arXiv:2306.11297}, 2023.

\bibitem{pujahari2022quantumreview2}
R.~M. Pujahari and A. Tanwar, ``Quantum federated learning for wireless communications,'' in \emph{Federated Learning for IoT Applications}, Springer, 2022, pp. 215--230.

\bibitem{chen2021federatedreview1}
S.~Y.-C. Chen and S. Yoo, ``Federated quantum machine learning,'' \emph{Entropy}, vol.~23, no.~4, p. 460, 2021.

\bibitem{chehimi2023foundations}
M.~Chehimi, S.~Y.-C.~Chen, W.~Saad, D.~Towsley, and M.~Debbah, 
``Foundations of quantum federated learning over classical and quantum networks,'' 
\emph{IEEE Network}, 2023.

\bibitem{ren2023towards}
C. Ren \emph{et al.}, ``Towards Quantum Federated Learning,'' \emph{arXiv preprint arXiv:2306.09912}, 2023.

\bibitem{larasati2022quantum}
H.~T. Larasati, M. Firdaus, and H. Kim, ``Quantum Federated Learning: Remarks and Challenges,'' in \emph{Proc. IEEE CSCloud/EdgeCom}, 2022, pp. 1--5.

\bibitem{yamany2021oqfl}
W. Yamany, N. Moustafa, and B. Turnbull, ``OQFL: An optimized quantum-based federated learning framework for defending against adversarial attacks in intelligent transportation systems,'' \emph{IEEE Transactions on Intelligent Transportation Systems}, 2021.

\bibitem{li2021quantumblind}
W. Li, S. Lu, and D.-L. Deng, ``Quantum federated learning through blind quantum computing,'' \emph{Science China Physics, Mechanics \& Astronomy}, vol.~64, no.~10, p. 100312, 2021.

\bibitem{xia2021quantumfed}
Q. Xia and Q. Li, ``Quantumfed: A federated learning framework for collaborative quantum training,'' in \emph{Proc. IEEE GLOBECOM}, 2021, pp. 1--6.

\bibitem{yang2021decentralizing}
C.-H.~H. Yang \emph{et al.}, ``Decentralizing feature extraction with quantum convolutional neural network for automatic speech recognition,'' in \emph{Proc. IEEE ICASSP}, 2021, pp. 6523--6527.

\bibitem{larasati2022quantumRemarks}
H.~T. Larasati, M. Firdaus, and H. Kim, ``Quantum Federated Learning: Remarks and Challenges,'' in \emph{Proc. IEEE CSCloud/EdgeCom}, 2022, pp. 1--5.

\bibitem{zhao2023nonQFLinf}
H. Zhao, ``Non-IID quantum federated learning with one-shot communication complexity,'' \emph{Quantum Machine Intelligence}, vol.~5, no.~1, p.~3, 2023.

\bibitem{chehimi2022quantum}
M. Chehimi and W. Saad, ``Quantum federated learning with quantum data,'' in \emph{Proc. IEEE ICASSP}, 2022, pp. 8617--8621.

\bibitem{xia2021defending}
Q. Xia, Z. Tao, and Q. Li, ``Defending against byzantine attacks in quantum federated learning,'' in \emph{Proc. IEEE MSN}, 2021, pp. 145--152.

\bibitem{yu2022quantum}
K. Yu, X. Zhang, Z. Ye, G.-D. Guo, and S. Lin, ``Quantum federated learning based on gradient descent,'' \emph{arXiv preprint arXiv:2212.12913}, 2022.

\bibitem{qi2023optimizingnaturalGD}
J. Qi, X.-L. Zhang, and J. Tejedor, ``Optimizing Quantum Federated Learning Based on Federated Quantum Natural Gradient Descent,'' in \emph{Proc. IEEE ICASSP}, 2023, pp. 1--5.

\bibitem{huang2022quantum}
R. Huang, X. Tan, and Q. Xu, ``Quantum federated learning with decentralized data,'' \emph{IEEE Journal of Selected Topics in Quantum Electronics}, vol.~28, no.~4, pp. 1--10, 2022.

\bibitem{moin2023model}
A. Moin, A. Badii, and M. Challenger, ``Model-Driven Quantum Federated Learning (QFL),'' \emph{arXiv e-prints}, p. arXiv:2304.XXXX, 2023.

\bibitem{xia2021defending}
Q. Xia, Z. Tao, and Q. Li, ``Defending against byzantine attacks in quantum federated learning,'' in \emph{Proc. IEEE MSN}, 2021, pp. 145--152.

\bibitem{narottama2023federated}
B. Narottama and S.~Y. Shin, ``Federated Quantum Neural Network with Quantum Teleportation for Resource Optimization in Future Wireless Communication,'' \emph{IEEE Transactions on Vehicular Technology}, 2023.

\bibitem{yun2022quantum}
W.~J. Yun, J.~P. Kim, H. Baek, S. Jung, J. Park, M. Bennis, and J. Kim, ``Quantum Federated Learning with Entanglement Controlled Circuits and Superposition Coding,'' \emph{arXiv preprint arXiv:2212.01732}, 2022.

\bibitem{zhang2022federated}
Y. Zhang, C. Zhang, C. Zhang, L. Fan, B. Zeng, and Q. Yang, ``Federated Learning with Quantum Secure Aggregation,'' \emph{arXiv preprint arXiv:2207.07444}, 2022.

\bibitem{ahmad2022dimensionalityDimensionReduction}
N. Ahmad and A.~B. Nassif, ``Dimensionality reduction: Challenges and solutions,'' in \emph{ITM Web of Conferences}, vol.~43, p.~01017, 2022.

\bibitem{barnard2019predictingDimensionReduction}
A.~S. Barnard and G. Opletal, ``Predicting structure/property relationships in multi-dimensional nanoparticle data using t-distributed stochastic neighbour embedding and machine learning,'' \emph{Nanoscale}, vol.~11, no.~48, pp. 23165--23172, 2019.

\bibitem{pokhrel2024quantum}
S.~R. Pokhrel, N. Yash, J. Kua, G. Li, and L. Pan, ``Quantum Federated Learning Experiments in the Cloud with Data Encoding,'' \emph{arXiv preprint arXiv:2405.00909}, 2024.

\bibitem{li2022federatedreimannian}
J. Li and S. Ma, ``Federated learning on Riemannian manifolds,'' \emph{arXiv preprint arXiv:2206.05668}, 2022.

\bibitem{huang2024riemannian}
Z. Huang, W. Huang, P. Jawanpuria, and B. Mishra, ``Riemannian Federated Learning via Averaging Gradient Stream,'' \emph{arXiv preprint arXiv:2409.07223}, 2024.

\bibitem{monga2021algounrolling}
V.~Monga, Y.~Li, and Y.~C. Eldar, ``Algorithm unrolling: Interpretable, efficient deep learning for signal and image processing,'' \emph{IEEE Signal Processing Magazine}, vol.~38, no.~2, pp.~18--44, 2021.

\bibitem{breast_cancer_wisconsin_1993}
W.~Wolberg, O.~Mangasarian, N.~Street, and W.~Street, ``Breast Cancer Wisconsin (Diagnostic),'' \emph{UCI Machine Learning Repository}, 1993. [Online]. Available: \url{https://doi.org/10.24432/C5DW2B}

\bibitem{panigrahi2023feddcs}
Monalisa Panigrahi, Sourabh Bharti, and Arun Sharma, 
``FedDCS: A distributed client selection framework for cross device federated learning,'' 
\textit{Future Generation Computer Systems}, vol.~144, pp.~24--36, 2023.


\bibitem{shen2025multiobjective}
Yuhao Shen, Wei Xi, Yunyun Cai, Yuwei Fan, He Yang, and Jizhong Zhao, 
``Multi-objective federated learning: Balancing global performance and individual fairness,'' 
\textit{Future Generation Computer Systems}, vol.~162, p.~107468, 2025.

\bibitem{yu2025towardsmultiobjective}
Guo Yu, Lianbo Ma, Xilu Wang, Wei Du, Wenli Du, and Yaochu Jin, 
``Towards fairness-aware multi-objective optimization,'' 
\textit{Complex \& Intelligent Systems}, vol.~11, no.~1, p.~50, 2025.

\bibitem{kang2024surveyperanto}
Shida Kang, Kaiwen Li, and Rui Wang, 
``A survey on Pareto front learning for multi-objective optimization,'' 
\textit{Journal of Membrane Computing}, pp.~1--7, 2024.

\bibitem{arai2024deep}
S.~Arai and S.~Takabe, 
``Deep Unfolded Local Quantum Annealing,'' 
\emph{arXiv preprint arXiv:2408.03026}, 2024.

\bibitem{thanasilp2022exponential}
S.~Thanasilp, S.~Cerezo, and P.~J.~Coles, 
``Exponential concentration and control of measurement noise in quantum circuits,'' 
\emph{arXiv:2210.10682}, 2022.

\bibitem{mitarai2018quantum}
K.~Mitarai, M.~Negoro, M.~Kitagawa, and K.~Fujii, 
``Quantum circuit learning,'' 
\emph{Physical Review A}, vol.~98, no.~3, p.~032309, 2018.


\end{thebibliography}
\end{document}